\documentclass[10.5pt,twocolumn,journal,letterpaper]{IEEEtran}

\pdfoutput=1 
\usepackage{amsmath, amsthm, amssymb}
\usepackage{url,flushend,multirow,booktabs}
\usepackage{algorithm,algorithmic}   % no crossline
\usepackage{graphicx}
\usepackage{bm}
\usepackage{cite}
\usepackage{subfigure}
\usepackage{float}
\usepackage{color}
\usepackage{multirow}
\usepackage[colorlinks,citecolor=green,urlcolor=blue,bookmarks=false,hypertexnames=true]{hyperref}
\usepackage{colortbl}
\definecolor{maroon}{cmyk}{0.08,0.04,0.00,0.06}  % light blue

\newcommand{\eg}{e.g.}
\newcommand{\ie}{i.e.}

\newcommand{\etal}{\textit{et al.}}

\DeclareFixedFont{\mf}{OT1}{ptm}{m}{n}{10pt}
\DeclareFixedFont{\mfb}{OT1}{ptm}{bx}{n}{10pt}

\begin{document}
%

\title{Residual Spatial Fusion Network for RGB-Thermal Semantic Segmentation}

\author{Ping~Li,~\IEEEmembership{Member,~IEEE}, Junjie~Chen, Binbin~Lin, and Xianghua~Xu 
%\thanks{Manuscript received May. 10th, 2023.}
\thanks{P.~Li, J.~Chen and X.~Xu are with the School of Computer Science and Technology, Hangzhou Dianzi University, Hangzhou, China (e-mail: lpcs@hdu.edu.cn, cjj@hdu.edu.cn, xhxu@hdu.edu.cn). P.~Li is also with Guangdong Laboratory of Artificial Intelligence and Digital Economy (SZ), China.}
\thanks{B.~Lin is with the College of Computer Science, Zhejiang University, Hangzhou, China. (e-mail:binbinlin@zju.edu.cn).} 
}
\markboth{Draft}
{LI \MakeLowercase{\textit{et al.}}:~Residual Spatial Fusion Network for RGB-Thermal Semantic Segmentation}

\maketitle

\begin{abstract}
  Semantic segmentation plays an important role in widespread applications such as autonomous driving and robotic sensing. Traditional methods mostly use RGB images which are heavily affected by lighting conditions, \eg, darkness. Recent studies show thermal images are robust to the night scenario as a compensating modality for segmentation. However, existing works either simply fuse RGB-Thermal (RGB-T) images or adopt the encoder with the same structure for both the RGB stream and the thermal stream, which neglects the modality difference in segmentation under varying lighting conditions. Therefore, this work proposes a Residual Spatial Fusion Network (RSFNet) for RGB-T semantic segmentation. Specifically, we employ an asymmetric encoder to learn the compensating features of the RGB and the thermal images. To effectively fuse the dual-modality features, we generate the pseudo-labels by saliency detection to supervise the feature learning, and develop the Residual Spatial Fusion (RSF) module with structural re-parameterization to learn more promising features by spatially fusing the cross-modality features. RSF employs a hierarchical feature fusion to aggregate multi-level features, and applies the spatial weights with the residual connection to adaptively control the multi-spectral feature fusion by the confidence gate. Extensive experiments were carried out on two benchmarks, \ie, MFNet database and PST900 database. The results have shown the state-of-the-art segmentation performance of our method, which achieves a good balance between accuracy and speed. 
 
\end{abstract}


 \ifCLASSOPTIONpeerreview
 \begin{center} \bfseries EDICS Category: 3-BBND \end{center}
 \fi

\IEEEpeerreviewmaketitle

\section{Introduction}
\label{sec1:intro}

\IEEEPARstart{S}{emantic} segmentation performs the pixel-level image classification, and it has wide applications, such as autonomous driving~\cite{caesar-cvpr2020-nuscenes}, video surveillance~\cite{collins-pami2000-introduction}, and scene understanding~\cite{zhou-ijcv2019-semantic}. Most existing works adopts the prevailing Fully Convolutional Neural networks (FCN) \cite{long-cvpr2015-fcn} by taking RGB images as the input. However, the resolution of RGB image is very sensitive to lighting conditions, \eg, its quality is satisfying under sufficient illumination in the daytime, and sharply deteriorates during nighttime, which poses a significant safety concern. Hence, recent works~\cite{ha-iros2017-mfnet} employ the thermal images to provide the coarse contours of objects under very poor lighting conditions for compensating the RGB images, named RGB-T semantic segmentation. This is because the gray-scale thermal image is generated by emitting the thermal radiation to the objects whose temperature are above the absolute zero, and most objects can be captured by the thermal image in adverse lighting conditions. 

\begin{figure*}[!t]
	\centering
	\includegraphics[width=1\linewidth]{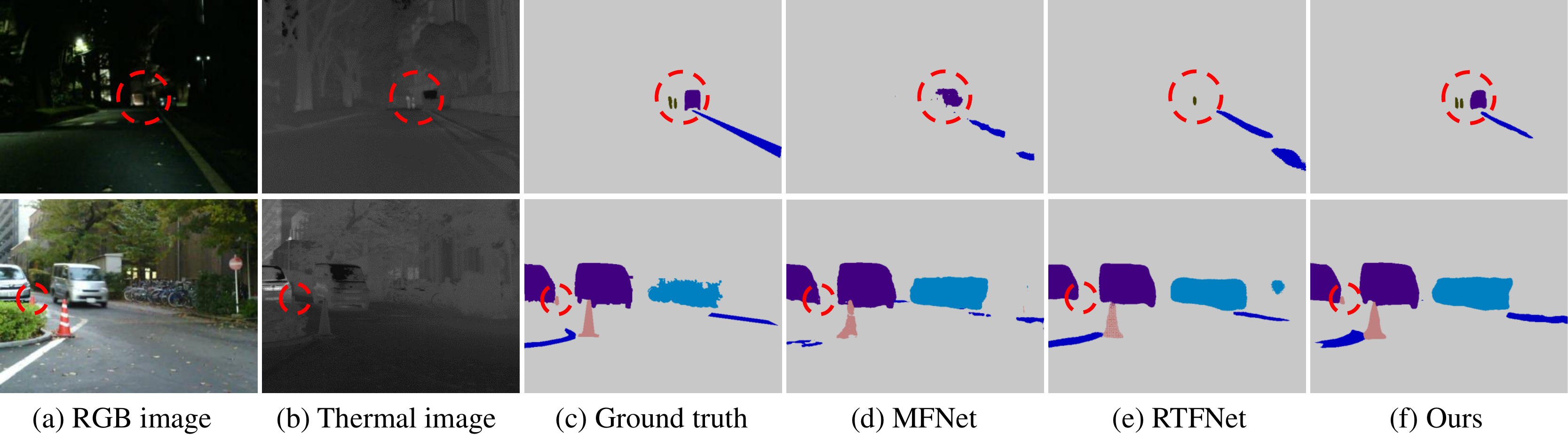}
	\caption{The RGB-T semantic segmentation results. The top row shows the nighttime image and the bottom row shows the daytime image. (a) RGB image; (b) Thermal image; (c) Ground truth; (d) MFNet~\cite{ha-iros2017-mfnet}; (e) RTFNet~\cite{sun-ral2019-rtfnet}; (f) Ours.}
	\label{fig:motivation}
	%\vspace{-1mm}
\end{figure*}

To this end, there has been much endeavor devoted to effectively learning the RGB and thermal features from the two compensating modalities. Previous works \cite{ha-iros2017-mfnet,sun-ral2019-rtfnet,sun-tase2021-fuseseg} often use the convolutional neural networks with the same structure as the backbone to extract their features, and adopt the indiscriminate fusion strategy to combine the intermediate feature maps by element-wise summation \cite{sun-ral2019-rtfnet,sun-tase2021-fuseseg} or concatenation \cite{ha-iros2017-mfnet}. However, such strategy fails to discriminate the different reliability of the two modalities under varying lighting conditions, \eg, thermal features are contaminated by RGB features. As depicted in Fig.~\ref{fig:motivation}, the top row shows the nighttime image where two persons and a car are surrounded by a dashed circle, and the bottom shows the daytime image where a traffic cone is marked by the circle. For the top image, the two persons are hard to see at night but thermal imaging camera can clearly capture them, so the thermal features are expected to be emphasized more in such situation. If adopting the indiscriminate fusion, the thermal features might be contaminated by the RGB features, leading to failure case. For the bottom image, the little traffic cone is not clearly seen from the thermal image since its heat temperature is very low, but it can be easily found in the RGB image. Under such circumstance, the RGB image should play the dominating role in segmentation, and the indiscriminate fusion might cause the confusion when fusing the two modalities. Therefore, it poses the main challenges for RGB-T semantic segmentation, \ie, how to effectively fuse the RGB and thermal features under different lighting conditions.

% ------------------------- Illustration of RGB-T Semantic Segmentation -------------------

To address this issue, inspired by \cite{chen-tip2020-dpanet}, we propose the Residual Spatial Fusion Network (RSFNet) for RGB-T semantic segmentation. Our method generates the pseudo-labels by the saliency detection on both RGB and thermal images, and derives the confidence score of the two modalities by regression. Then, their features are adaptively fused by a gate mechanism. To obtain the pseudo-labels of the image, we adopt the saliency detection \cite{montabone-ivc2010-human} skill to yield the binary saliency map, which reflects the discriminant ability of the multi-spectral image. And the IoU score of the binary saliency map and the ground-truth label is regarded as the pseudo-label, which is expected to be close to the confidence score derived by regression, \ie, feeding the feature map to the fully-connected (fc) layers followed by the sigmoid function. When the IoU score is high, it means the saliency area might be more possibly the target area. Both RGB encoder and thermal encoder interact with each other by the Residual Spatial Fusion (RSF) module during each corresponding stage of the encoder (\eg, ResNet~ \cite{he-cvpr2016-resnet}). RSF module takes the RGB and thermal feature maps and their confidence scores as the input, while it outputs the enhanced features of each layer to the decoder to derive the final label map. Here, the RGB and thermal confidence scores act as the gate value to control the multi-spectral feature information passed to the fusion module, and the more contributing features under the present lighting condition should be better preserved compared to the less informative ones. 

To further improve the segmentation performance of our RSFNet, we adopt the structural re-parameterization \cite{ding-cvpr2021-repvgg, ding-cvpr2021-dbb} strategy for the RSF module without additional inference time cost. This strategy decouples the multi-branch training and single-branch inference, \ie, the multi-branch structure parameters are transformed to single-branch structure parameters. In this way, it not only guarantees the fast single-branch inference speed and low memory cost, but also enjoys the high performance of multi-branch structure. Different from previous works~\cite{ding-cvpr2021-repvgg,ding-cvpr2021-dbb}, we develop the factorized convolution~\cite{szegedy-cvpr2016-inceptionv3} for concatenation in the fusion module to capture the multi-scale features. 

The main contributions are summarized as follows:
\begin{itemize}
  \item We propose the Residual Spatial Fusion module to take advantage of the two compensating modalities, \ie, RGB image and thermal image. Specifically, the feature learning in both RGB branch and thermal branch is guided by the other modality at each ResNet stage in the encoder. Moreover, the structural re-parameterization strategy is applied to the fusion module to capture better cross-modality features.

  \item We adopt an asymmetric structure to encode the RGB images by the larger backbone (\eg, ResNet101) and the thermal images by the smaller backbone (\eg, ResNet34), which reduces the computational cost, allowing the model to be readily deployed in real-time applications.
  
  \item We design the Pseudo-Label Generation module to yield the pseudo-labels for supervising the RGB and the thermal feature learning, and also to help better fuse the RGB-T features.
  
  \item Extensive experiments were carried out on two benchmarks, including MFNet dataset~\cite{ha-iros2017-mfnet} and PST900 dataset~\cite{shivakumar-icra2020-pst900}. Both quantitative and qualitative results have demonstrated the superiority of the proposed framework.

\end{itemize}

The rest of this paper is organized as follows. Section~\ref{related} reviews some closely related works and Section~\ref{method} introduces our Residual Spatial Fusion Network. Then, we report the experimental results on two benchmarks to demonstrate the advantage of our method in Section~\ref{test}. Finally, we conclude this work in Section~\ref{conclusion}.
%

%-------------------------------------------------------------------------
\section{Related Work}
\label{related}
Recent years have witnessed remarkable progress in semantic segmentation. In the following, we briefly discuss the related works of RGB semantic segmentation and RGB-T semantic segmentation.

\subsection{RGB Semantic Segmentation} 
Traditionally, semantic segmentation takes RGB images as the input and predicts the pixel-level labels. Early works adopt the hand-crafted features and use the classical classifier for segmentation, such as Boosting~\cite{tu-pami2009-auto}, Support Vector Machine~\cite{fulkerson-iccv2009-class}, and Random Forest~\cite{shotton-cvpr2008-semantic}, and later works employ the rich context~\cite{carreira-eccv2012-semantic} and structural prediction~\cite{carreira-pami2011-cpmc} to improve segmentation performance. However, the performance is still restricted by the limited representation ability of those image features. In the last decade, Convolutional Neural Networks (CNNs) have shown the strong power in learning better features for image classification, which lead to the performance breakthrough. This inspires Long \etal~\cite{long-cvpr2015-fcn} to develop Fully-Convolution Network (FCN), which establishes a milestone in semantic segmentation.    

Recent semantic segmentation works are mostly developed upon FCN, \eg, Chen's group~\cite{chen-pami2017-deeplabv2, chen-arxiv2017-deeplabv3, chen-eccv2018-deeplabv3+} proposed a series of Deeplab models that use the dilated convolution to build the Atrous Spatial Pyramid Pooling (ASPP) module to enlarge the receptive field, so as to capture the multi-scale contextual feature. Similarly, Zhao \etal~\cite{zhao-cvpr2017-pspnet} built the pyramid pooling module to segment the objects with different sizes. Nevertheless, these methods only locally model the spatial context and fail to encode the global context. To handle this drawback, Wang \etal~\cite{wang-cvpr2018-nonlocal} designed the non-local block that adopting the self-attention mechanism~\cite{vaswani-nips2017-attention} to capture the long-range dependency among the pixels in image, and Fu \etal~\cite{fu-cvpr2019-danet} considered the self-attention at both pixel level and channel level. Moreover, Yu \etal~\cite{yu-eccv2018-bisenet} designed a dual-path encoder that include the spatial path that yields the high-definition feature map and the context path that captures the contextual relation using fast down-sampling, on the basis of which Yu \etal~\cite{yu-ijcv2021-bisenetv2} developed a lighter module to learn features and employed the guided aggregation layer to enhance the fusion of dual-path features. Unlike attention-based methods that model the pixel-wise similarity, Liu \etal~\cite{liu-cvpr2022-caanet} adopted the Class Activation Map (CAM) to model pixel relations. 

To capture the spatial long-range dependency, there have emerged some sequence-to-sequence methods that employ the self-attention based Vision Transformer (ViT)~\cite{dosovitskiy-arxiv2020-vit} to derive features. For example, Zheng \etal~\cite{zheng-cvpr2021-setr} used the vanilla ViT as the encoder to obtain the hierarchical feature maps, which are sent to the decoder that adopts the progressive upsampling with multi-level feature aggregation to obtain segmentation feature maps; Wang \etal~\cite{wang-aaai2022-uctransnet} proposed the channel-wise cross fusion transformer to fuse the multi-scale context along the channel dimension, which benefits bridging the semantic gap between the encoder and the decoder; Zhang \etal~\cite{zhang-cvpr2022topformer} utilized the advantages of both CNN and ViT, by feeding the multi-scale image token to ViT to derive the global scale-aware semantics, which are then fused with the local tokens to achieve the more accurate segmentation map.

However, the above semantic segmentation methods only take the RGB images as the input, which do not work well under poor lighting conditions, such as urban scenes at night. So we consider the thermal images as the compensating modality to join the segmentation framework with RGB images for better segmentation.

\subsection{RGB-Thermal Semantic Segmentation} 
To compensate the insufficient vision information from RGB images, recent works \cite{ha-iros2017-mfnet, shivakumar-icra2020-pst900, fan-tnnls2020-rethinking} attempted to use the multi-spectral camera to take multi-spectral photos, including RGB-Thermal image and RGB-Depth image \cite{zhu-tmm2023-s3net}, which are often used in saliency detection \cite{liu-tmm2020-fusion, cong-tmm2022-thermal, tu-tmm2022-adfnet} and tracking \cite{xu-tmm2021-multimodal}. This work concentrates on the RGB-T semantic segmentation, which takes the paired RGB and thermal images as the input, and predicts the class label of each pixel in a given image. 

Existing RGB-T semantic segmentation works \cite{ha-iros2017-mfnet, sun-ral2019-rtfnet, sun-tase2021-fuseseg} often use two symmetric encoders (\ie, the same backbone) to extract both RGB and thermal features, and fuse the multi-spectral features indistinguishably. For example, Ha \etal~\cite{ha-iros2017-mfnet} proposed the Multi-spectral Fusion Networks (MFNet), which adopts the VGG-like network \cite{simonyan-arxiv2014-vgg} as the encoder for both RGB and thermal images, and then concatenate their features. Similarly, Sun \cite{sun-ral2019-rtfnet} adopted two ResNet152 \cite{he-cvpr2016-resnet} as the encoder. Later, Sun \cite{sun-tase2021-fuseseg} employed two DenseNet161 \cite{huang-cvpr2017-densenet} as the encoder, and the RGB-T features are fused by element-wise summation. Similarly, Zhou \cite{zhou-tmm2021-mffenet} utilized two DenseNet121 as the encoder whose features are also fused by element-wise summation, and then used the compact ASPP \cite{chen-eccv2018-deeplabv3+} module to obtain multi-scale feature. Such kind of indistinguishable fusion strategy neglects the different reliability of two modalities, and one modality may be contaminated by the other one under poor lighting conditions. To circumvent this situation, Zhang \cite{zhang-cvpr2021-abmdrnet} adopted the bridging-then-fusing strategy to consider the modality difference, and fused the multi-spectral feature fusion along the spatial and the channel dimensions. But its multi-scale spatial context module requires intensive computing sources due to the self-attention mechanism.

In addition, Zhou \cite{zhou-tip2021-gmnet} divided the features to junior-, intermediate-, and senior-level, and obtained the different-level features using both shallow and deep feature fusion schemes to focus on different areas in an image. They \cite{zhou-aaai2022-egfnet} also tried to use the prior edge information to help capture the object contours in RGB-T images, and embedded the edge-aware prediction map to the feature maps to find more details. Furthermore, Zhou \cite{zhou-tiv2022-mtanet} proposed the multitask-aware network (MTANet) with hierarchical multimodal fusion, and built a high-level semantic module to help emerge the coarse features at different abstraction levels. 

However, the above RGB-T semantic segmentation methods adopt the symmetric encoders for both RGB and thermal images, which might be improper because the thermal image only takes up a single channel. So we argue that the large backbone with more layers is more suitable for RGB images, while the the smaller one is better for thermal images at the reduced cost.

%-------------------------------------------------------------------------
\section{The Proposed Method}
\label{method}
\begin{figure*}[!t]
	\centering
	\includegraphics[width=0.9\textwidth]{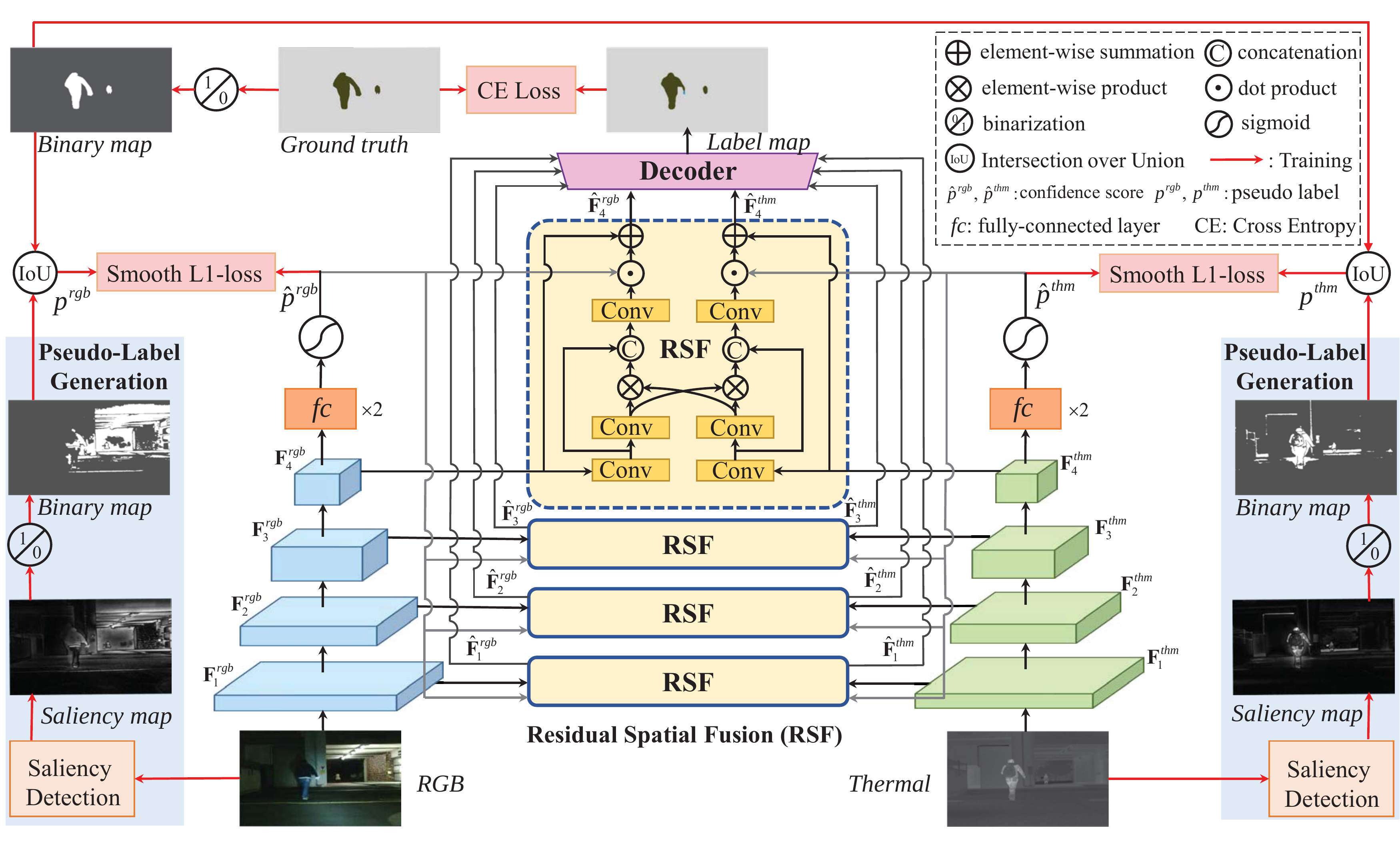}
	\caption{Overview of the proposed RSFNet framework. It consists of the encoder (the inner RGB branch and thermal branch), Residual Spatial Fusion (RSF) module (in the middle), decoder, and the pseudo-label generation module (the left-most RGB branch and the right-most thermal branch). A pair of RGB image and thermal image are fed to the asymmetric encoder to derive the confidence scores $\{\hat{p}^{rgb}, \hat{p}^{thm}\}$ by regression, and they are simultaneously input to the Pseudo-Label Generation (PLG) module to obtain the binary saliency map for computing the IoU scores $\{p^{rgb}, p^{thm}\}$ (soft labels), which are regarded as the pseudo-labels to supervise feature learning. The RSF module accepts both the confidence score and the intermediate feature map, yielding the cross-modal feature maps, which are then decoded to the label map as the final prediction.}
	\label{fig:framework}
	%\vspace{-1mm}
\end{figure*}

This section describes the proposed RSFNet framework as illustrated in Fig.~\ref{fig:framework}, which mainly includes asymmetric encoder, Pseudo-Label Generation (PLG), Residual Spatial Fusion (RSF), and decoder. We firstly show the problem definition of the task.

% -------------- RSFNet Framework -----------------

\subsection{Problem Definition}
The RGB-T semantic segmentation task aims to predict the pixel-level labels given a pair of RGB image and thermal image. Mathematically, the RGB-T image is concatenated by RGB image $\mathbf{I}\in \mathbb{R}^{3\times H\times W} $ and thermal image $\mathbf{I}^{thm}\in\mathbb{R}^{1\times H\times W}$, denoted by a four-channel tensor $\mathbf{I}= [\mathbf{I}^{rgb}; \mathbf{I}^{thm}]\in\mathbb{R}^{4\times H \times  W} $, where $H$ is the height and $W$ is the width; and its ground-truth label is $\mathbf{Y}\in \mathbb{R}^{C\times{H}\times{W}}$, where $C$ is the number of semantic classes. Our goal is to learn a segmentation model that effectively fuses both RGB and thermal features, which outputs the label map of an image for fast inference time.  

\subsection{Asymmetric Encoder}
As mentioned earlier, most existing works \cite{sun-ral2019-rtfnet,sun-tase2021-fuseseg} adopt the symmetric encoder, \ie, learn RGB and thermal features using the same backbone regardless of the modality difference. For example, RTFNet \cite{sun-ral2019-rtfnet} employs the large backbone ResNet-152 \cite{he-cvpr2016-resnet} to learn the features of two modalities indistinguishably. Generally, RGB images provide rich color and texture information using three channels, while thermal images indicate coarse location and obscure appearance using a single channel. Hence, we consider to discriminate the feature learning using the asymmetric encoder, \ie, the larger backbone (\eg, ResNet101) for the RGB branch and the smaller backbone (\eg, ResNet34) for the thermal branch. In this way, the computational costs can be largely reduced compared to the symmetric encoder.

Take ResNet101 for example, it has four stages, indexed by $s\in \{1, 2, 3, 4\}$, and each stage outputs the RGB feature map $\mathbf{F}^{rgb}_s \in \mathbb{R}^{C^{rgb}_s\times{H_s}\times{W_s}}$ and the thermal feature map $\mathbf{F}^{thm}_s \in \mathbb{R}^{C^{thm}_s\times{H_s}\times{W_s}}$. Here, $\{H_1, W_1\} = \{\frac{H}{4}, \frac{W}{4}\}$, $\{C_1^{rgb}, C_1^{thm}\} = \{256, 256\}$, and $\{H_{s+1}, W_{s+1}, C_{s+1}^{rgb}, C_{s+1}^{thm}\} = \{\frac{H_s}{2}, \frac{W_s}{2}, 2 C_s^{rgb}, 2 C_s^{thm}\}$ if $s>1$. Note that $\{C_1^{rgb}, C_1^{thm}\} = \{64, 64\}$ for ResNet34.

\subsection{Pseudo-Label Generation}
To supervise the feature learning, we design a Pseudo-Label Generation (PLG) module that uses the saliency detection skill to derive the saliency map of RGB image and thermal image. The saliency map reflects the most salient area of the image according to the biology theory\cite{itti-pami1998-model}, and such area probably overlaps with the target semantic area. Hence, we treat the IoU score of the binary saliency map and the binary ground-truth as the soft label, termed pseudo-label. The pseudo-label is used for guiding the feature learning by computing its loss with the confidence score derived from the regression in the encoder, such that the learned feature map might focus more on the saliency area. Here, the confidence indicates the possibility of the learned RGB or thermal features capturing the overlap of the saliency map and the ground-truth label, which are available only at the training phase. During inference, the pseudo-label is unavailable, and the confidence score as some gate plays an important role in guiding the fusion of the dual-modality features by strengthening the high-confidence feature and weakening the low-confidence feature.

Firstly, we adopt the fine-grained saliency detection skill \cite{montabone-ivc2010-human} to compute the saliency maps $\{\mathbf{M}^{rgb}, \mathbf{M}^{thm}\} \in\mathbb{R}^{H\times W}$ for the paired RGB image and thermal image. For brevity, we omit the uppercase letters ``rgb'' and ``thm'' in the following. Second, we use the Ostu binarization approach \cite{otsu-tsmc1979-ostu} to obtain the binary saliency map $\tilde{\mathbf{M}} \in\mathbb{R}^{H\times W}$. Third, we evaluate the relevance of the binary saliency map and the ground-truth label map $\mathbf{Y}\in \mathbb{R}^{C\times{H}\times{W}}$. In particular, we separate the labeled foreground from the unlabeled background by binarization and obtain the binary ground-truth label map $\tilde{\mathbf{Y}}\in\mathbb{R}^{H\times W}$, and then compute the IoU (Intersection over Union) score of the binary saliency map and the binary ground-truth label map, \ie,
\begin{equation} \label{eq:iou}
	IoU = \frac{| \tilde{\mathbf{M}} \cap \tilde{\mathbf{Y}} |}{| \tilde{\mathbf{M}} \cup \tilde{\mathbf{Y}} |}.
\end{equation}
Now we define the IoU score as the pseudo-label for the feature learning in the encoder. And we have $\{p^{rgb}, p^{thm}\}$ for the RGB image and the thermal image, and their values indicate whether the RGB or the thermal saliency map is more relevant to the ground-truth label map.

Essentially, the generated pseudo-label is utilized to supervise the feature learning by guiding the regression of the confidence score. And the confidence score suggests to what degree the RGB or the thermal feature map captures the saliency area of the image by feature encoding. Specifically, the Global Average Pooling $GAP(\cdot)$ is operated on the output feature map of the encoder branch, and then two fully-connected layers $fc(\cdot)$ as well as the sigmoid function $\sigma(\cdot)$ are followed to yield the confidence score $\hat{p} = \sigma(fc(GAP(\mathbf{F}_4)) $, \ie, the estimated value in the regression. And we have $\{\hat{p}^{rgb}, \hat{p}^{thm}\}$ for the RGB image and the thermal image, and their values indicate whether the RGB or the thermal features dominate the saliency area of the input image. 

% -------------- RSF module -----------------
\begin{figure}[!t]
	\centering
	\includegraphics[width=0.3\textwidth]{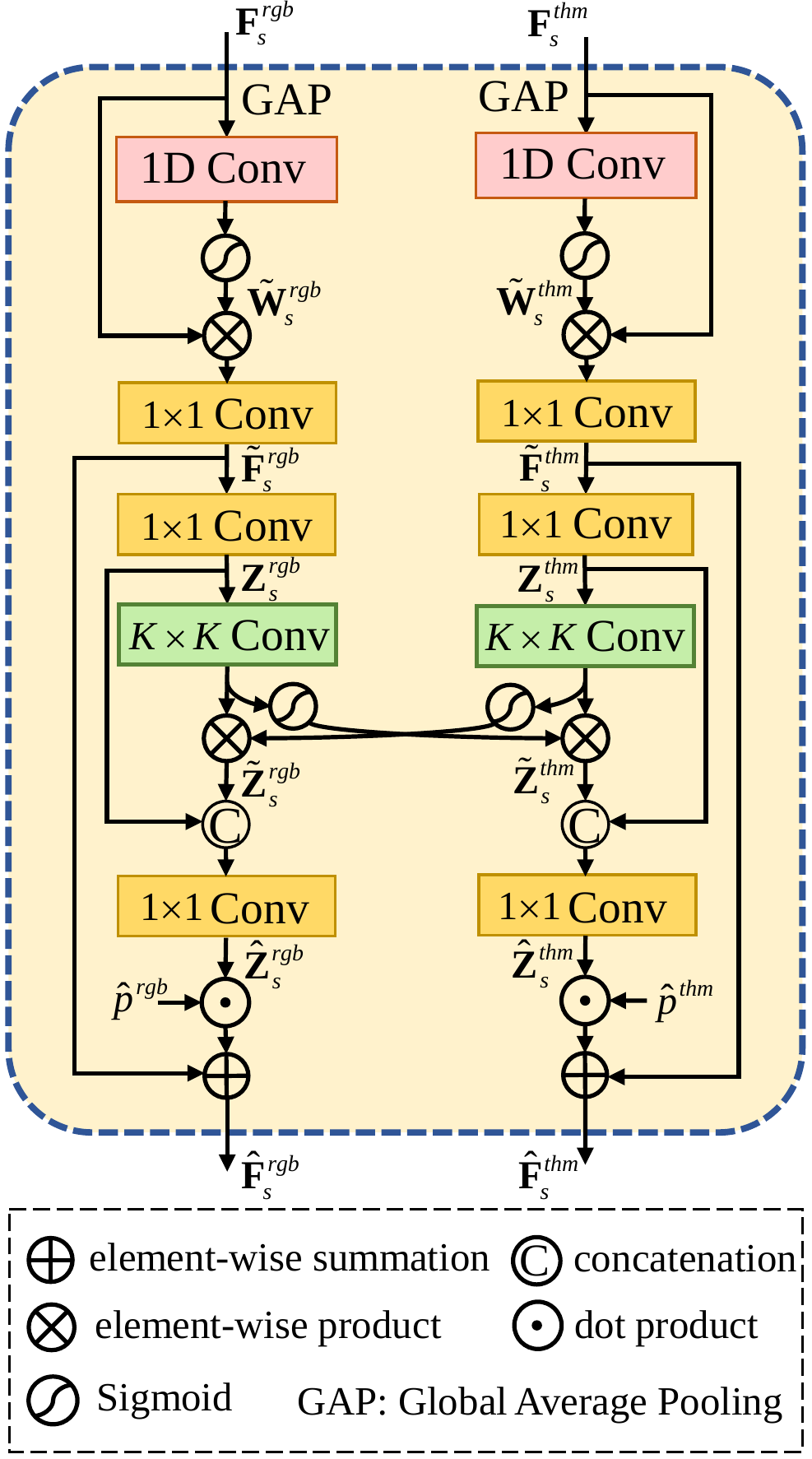}
	\caption{The architecture of the Residual Spatial Fusion (RSF) module.}
	\label{fig:fusion_module}
	%\vspace{-1mm}
\end{figure}

\subsection{Residual Spatial Fusion}
To alleviate the modality contamination problem, \ie, the thermal feature is affected by the RGB feature under poor lighting condition, we proposed the Residual Spatial Fusion (RSF) module, as illustrated in Fig.~\ref{fig:fusion_module}. This module takes the RGB and thermal confidence scores $\{\hat{p}^{rgb}, \hat{p}^{thm} \}$ as the gate to control the fusion ratio for the cross-modal features. In principle, the features derived from the encoder with the higher confidence score should contribute more to the segmentation under varying lighting conditions. The details and the design idea are described below.

\textbf{Feature Recalibration}. As we know, the channel dimension of the RGB feature map is much more than that of the thermal feature map, since the model adopts the asymmetric encoder, \ie, the former uses the much larger backbone than the latter. Hence, the channel dimensions of the two modalities should be changed to the same before the fusion. It is common to use $1\times 1$ convolution (\ie, 2D convolution) to reduce the channel dimension to the same, but the pure $1\times 1$ convolution might cause the missing of some important attributes \cite{huang-iccv2021-fapn}. To handle this issue, instead of using the costly channel-attention \cite{hu-cvpr2018-senet} and self-attention \cite{vaswani-nips2017-attention}, inspired by \cite{wang-cvpr2020-eca, huang-iccv2021-fapn}, we employ the fast 1D (one-dimensional) convolution $Conv1D(\cdot)$ to do feature recalibration (normalizing the channel weights), which is more efficient, such that the important feature channels are emphasized while the redundant channels are suppressed. This results in the channel weight, \ie,
\begin{equation}
	\tilde{\mathbf{W}}_s = \sigma(Conv1D(GAP(\mathbf{F}_s)) \in \mathbb{R}^{C_s\times 1\times 1},
\end{equation}
which reflects the importance of each channel. The channel weight is used to do element-wise product ``$\otimes$'' with the feature map $\mathbf{F}_s$, and then pass the weighted feature map to the $1\times 1$ convolution block $Conv2D(\cdot)$ for reducing the channel dimension, leading to the low-dimensional feature, \ie
\begin{equation}
	\tilde{\mathbf{F}}_s = Conv2D(\tilde{\mathbf{W}}_s \otimes{\mathbf{F}_s}) \in \mathbb{R}^{\tilde{C}_s\times H_s\times W_s},
\end{equation}	
where $\tilde{C}_s \ll C_s$. Here, we set $\{\tilde{C}_1, \tilde{C}_2, \tilde{C}_3, \tilde{C}_4 \} = \{64, 128, 256,256\}$. 

\textbf{Cross-modality Fusion}. To fuse the RGB and the thermal features, inspired by the bottleneck block \cite{he-cvpr2016-resnet}, we design the light residual spatial fusion scheme to fully utilize the features of the two compensating modalities. Both of the feature maps are fed to a sequence of three 2D convolutions with the kernel size of $1\times 1$, $K\times K$, and $1\times 1$. The first convolution reduces the feature dimension by computing
\begin{equation}
  \mathbf{Z}_s = ReLU(BN(Conv2D(\tilde{\mathbf{F}}_s)))\in \mathbb{R}^{\tilde{C}\times H_s \times W_s},
\end{equation}
where the reduced dimension $\tilde{C}$ is set to 64, $BN(\cdot)$ denotes the batch normalization, and $ReLU(\cdot)$ is the activation function to add the non-linearity of the feature. The middle convolution followed by the sigmoid function $\sigma(\cdot)$ and the element-wise product $\otimes$ is used to do the cross spatial fusion of the low-dimensional dual-modality features, obtaining the cross-modality feature $\tilde{\mathbf{Z}}_s \in \mathbb{R}^{\tilde{C}\times H_s \times W_s}$. Such fusion employs the other modality feature as the weight map along the spatial dimension, which models the long-range pixel-wise relations of image. Meanwhile, the cross-modality feature is concatenated with low-dimensional feature $\mathbf{Z}_s$ by the residual connection along the channel dimension, \ie, $[\tilde{\mathbf{Z}}_s; \mathbf{Z}_s]$, which is then passed to the third convolution to recover the channel dimension to $\tilde{C}_s$, leading to the feature map $\hat{\mathbf{Z}}_s \in \mathbb{R}^{\tilde{C}_s\times H_s\times W_s}$.

\textbf{Fusion Gate}. As mentioned earlier, the confidence score $\hat{p}$ that reflects how well the RGB feature or the the thermal feature captures the saliency area of the image, act as the fusion gate to filter out the redundancy of the cross-modality feature $\hat{\mathbf{Z}}_s$. In this way, the feature of the modality with higher confidence receives more attention while that with lower confidence receives less attention during the dual-modality fusion. Specifically, the enhanced feature map of the $s$-th stage is denoted by $\hat{\mathbf{F}}_s = \tilde{\mathbf{F}}_s + \hat{p}\cdot \hat{\mathbf{Z}}_s \in \mathbb{R}^{\tilde{C}_s\times H_s\times W_s}$.   

% -------------- Structural Re-parameterization -----------------
\begin{figure}[!t]
	\centering
	\includegraphics[width=0.25\textwidth]{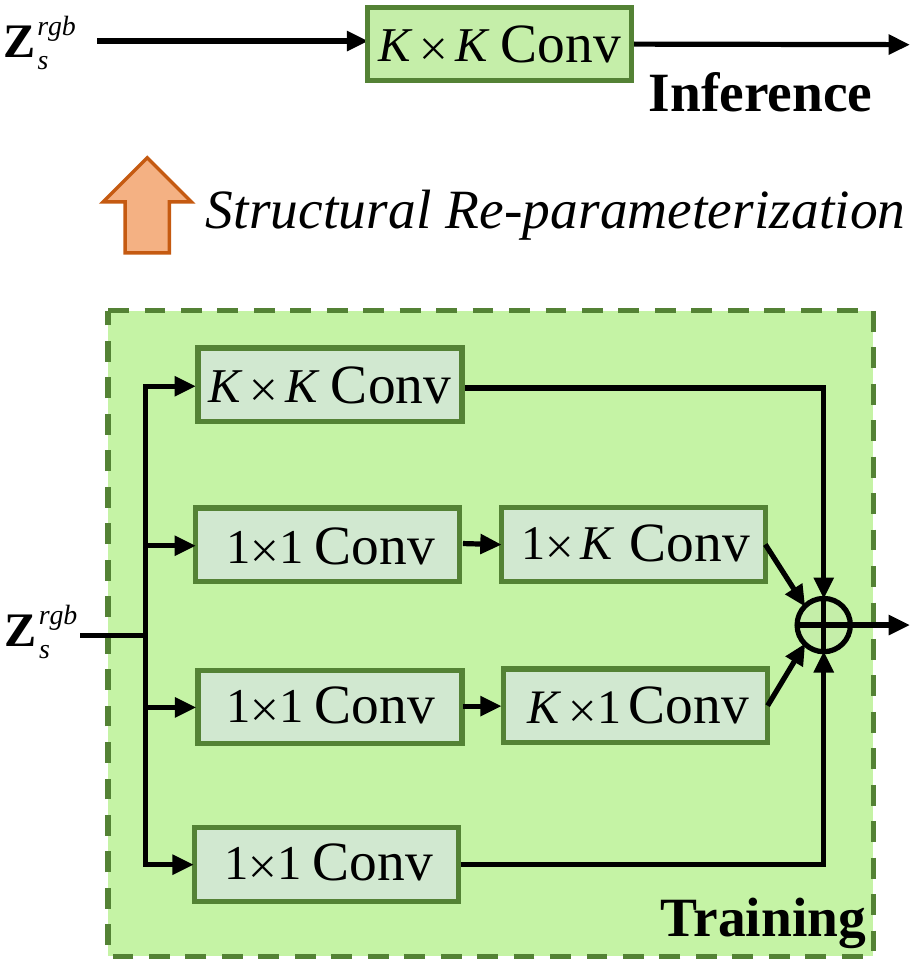}
	\caption{The architecture of the Residual Spatial Fusion (RSF) module.}
	\label{fig:repara}
	%\vspace{-1mm}
\end{figure}

\textbf{Structural Re-parameterization}. To make the RSF module capture the multi-scale feature, we adopt the structural re-parameterization \cite{ding-cvpr2021-repvgg, ding-cvpr2021-dbb} skill to improve the fusion performance with no more inference cost. As shown in Fig.~\ref{fig:repara}, for the $K\times K$ convolution block, we use the multi-branch fusion during training, and the original single branch during inference by re-parameterization. This ensures the fast inference at a lower memory cost when deploying the single branch fusion module in highly-demanding applications. In detail, we enlarge the $K\times K$ convolution by adding the $1\times 1$ convolution and two series of convolution groups, \ie, the convolution kernel size of $1\times K$ and $K\times 1$ with $1\times 1$ convolution, such that the multi-branch module enables to model the multi-scale cross-modality feature. At the inference period, all paralleled convolutions are transformed to a single $K\times K$ convolution by adopting the parameter transformation \cite{ding-cvpr2021-repvgg}. 

For example, the input of the $K\times K$ convolution block is $\mathbf{Z}_s \in \mathbb{R}^{\tilde{C}_s\times H_s\times W_s}$, and its parameter tensor is $\mathcal{K}\in \mathbb{R}^{\tilde{C}\times \tilde{C}\times K \times K}$. For the $1\times 1$ and $1\times K$ convolution block, the corresponding parameter tensors are $\mathcal{K}_1\in \mathbb{R}^{\tilde{C}\times \tilde{C}\times 1 \times 1}$ and $\mathcal{K}_2\in \mathbb{R}^{\tilde{C}\times \tilde{C}\times 1 \times K}$. If we feed the feature map $\mathbf{Z}_s$ to this block, it outputs the feature with the same size, \ie, 
\begin{equation}
	\mathbf{O}_s = \mathbf{Z}_s\ast \mathcal{K}_1 \ast \mathcal{K}_2\in \mathbb{R}^{\tilde{C}_s\times H_s\times W_s},
\end{equation}
where ``$\ast$'' denotes the convolution operator. According to \cite{ding-cvpr2021-dbb}, $1\times 1$ convolution is also known as the point-wise convolution \cite{howard-arxiv2017-mobilenets}, which does the channel-wise linear combination rather than the spatial-wise aggregation. So one can use the linear recombination to reshape the $1\times 1$ and $1\times K$ convolutions by combining their parameters, leading to the new $1\times K$ convolution, \ie, $\mathcal{K}^\prime = \mathcal{K}_2\ast \mathcal{K}_1^\top\in \mathbb{R}^{\tilde{C}\times \tilde{C}\times 1 \times K}$, where ``$(\cdot)^\top$'' denotes the tensor transpose operator. By zero-padding, the new convolution is reshaped to $K\times K$ convolution, \ie, $\hat{\mathcal{K}}\in \mathbb{R}^{\tilde{C}\times \tilde{C}\times K \times K}$, whose entries are assigned to the original $\mathcal{K}$. 

\subsection{Decoder}
We use the two-stream decoder to yield the predicted label map $\hat{\mathbf{Y}} \in \mathbb{R}^{C\times H\times W}$ by upsampling and fusing the enhanced feature maps $\{\hat{\mathbf{F}}_s^{rgb}, \hat{\mathbf{F}}_s^{thm}\}$. Both the RGB stream and the thermal stream include three blocks, each of which is comprised of three $3\times 3$ convolutions and two bilinear interpolations. And we adopt the multi-scale fusion \cite{chen-tip2020-dpanet} to consider both the high-level context and low-level details of the features. Note that the sigmoid function is used to normalize the predicted label values to $[0,1]$.

\subsection{Loss Function}
The total loss $\mathcal{L}$ of the proposed RSFNet consists of two parts, \ie, segmentation loss $\mathcal{L}_{seg}$ and regression loss $\mathcal{L}_{reg} = \mathcal{L}_{reg}^{rgb} + \mathcal{L}_{reg}^{thm}$. Formally, it is expressed by
\begin{equation}
	\mathcal{L} = \mathcal{L}_{seg} + \lambda \mathcal{L}_{reg},
\end{equation}
where the constant $\lambda>0$ (set to 0.3) balances the contribution of the regression loss to the model training. The former targets accurate segmentation predictions, and the latter focuses on learning the features to capture the saliency area.

\textbf{Segmentation Loss}. We adopt the commonly used multi-class cross-entropy loss to evaluate the error between the predicted label $\hat{\mathbf{Y}}$ and the ground-truth one $\mathbf{Y}$. Due to the class imbalance problem in RGB-T segmentation task, we follow \cite{ha-iros2017-mfnet} to use the weighted version, \ie,
\begin{equation} \label{eq:loss_seg}
	\mathcal{L}_{seg} = - \frac{1}{H\times W} \sum^{H}_{h=1}\sum^{W}_{w=1} \omega_{h,w} \mathbf{Y}_{h,w} \log \hat{\mathbf{Y}}_{h,w},
\end{equation}
where $(h, w)$ denotes the pixel coordinate, $\omega_{h,w}$ is the pixel weight computed as indicated in \cite{paszke-arxiv2016-enet}, \ie, $\omega_c = \frac{1}{\ln (1.05+p_c)}$, where the lowercase letter $c \in \{1, 2, \cdots, C\}$ is the class index, and $p_c$ is the class probability in training data.

\textbf{Regression Loss}. To encourage the learned feature map to capture the saliency area, we adopt the smooth L1-loss \cite{girshick-iccv2015-rcnn} to make the confidence score regress to the pseudo label. For the RGB image and the thermal image, it is respectively defined as
\begin{equation}
\mathcal{L}^{rgb}_{reg} = 
\left\{
\begin{array}{ll}
	0.5(e^{rgb})^2, & \text{if}~ e^{rgb} < 1, \\
	e^{rgb} - 0.5, & \text{otherwise}.
\end{array}
\right.
\end{equation}
and
\begin{equation}
	\mathcal{L}^{rgb}_{thm} = 
	\left\{
	\begin{array}{ll}
		0.5(e^{thm})^2, & \text{if}~e^{thm} < 1, \\
		e^{thm} - 0.5, & \text{otherwise}.
	\end{array}
	\right.
\end{equation}
where $e^{rgb} = |p^{rgb}-\hat{p}^{rgb}|$ and $e^{thm} = |p^{thm}-\hat{p}^{thm}|$ denote the absolute values of the error between the estimated confidence and the pseudo label value.

%-------------------------------------------------------------------------
\section{Experiment}
\label{test}
This section shows extensive experiments of semantic segmentation on two benchmark data sets. All experiments were conducted on a machine with a single GeForce RTX 2080Ti graphics card (11GB memory), and our RSFNet model was compiled using PyTorch 1.8, Python 3.8, and CUDA 10.2.

%-------------------------------------------------------------------------
\subsection{Datasets and Evaluation Metrics}
\textbf{MFNet dataset}\footnote{https://www.mi.t.u-tokyo.ac.jp/static/projects/mil\_multi-spectral} \cite{ha-iros2017-mfnet}. This is the first RGB-Thermal image dataset, which contains 1,569 urban scene images (820 at daytime and 749 at nighttime) with the 480$\times$640 spatial resolution. These RGB-T images are taken by an InfRec R500 camera, which captures images in both the visible and thermal infrared spectrum with different lens and sensors. They are categorized into eight semantic classes and one background class, including car, person, bike, curve, car stop, guardrail, color cone, and bump. The ratio of the training, the validation, and the test set is 2:1:1 for both the daytime and the nighttime images. Note that the proportion of each class is highly imbalanced, \eg, the guardrail images only take up 0.095\%.

\textbf{PST900 dataset}\footnote{https://github.com/ShreyasSkandanS/pst900\_thermal\_rgb} \cite{shivakumar-icra2020-pst900}. This is second RGB-Thermal image dataset, which has 894 synchronized and calibrated RGB and thermal image pairs with pixel-level human annotations across four semantic classes and one background class, involving fire extinguisher, backpack, hand-drill, and survivor. These images are taken by the FLIR Boson 320 and Stereolabs ZED Mini 3D cameras installed in a quadruped mobile robot platform from a variety of challenging subterranean environments, such as tunnels, mines, and caves. The ratio of the training and the test set is 2:1.

\textbf{Evaluation Metrics}. Following previous works \cite{shivakumar-icra2020-pst900}, we adopt two commonly used metrics, \ie, mean Accuracy (mAcc) and mean IoU (mIoU), to evaluate the semantic segmentation performance of the compared methods. The former is the average accuracy over all classes, and the latter is the average IoU score of the predicted and the ground-truth semantic region across all classes.

\subsection{Experimental Settings}
\textbf{Backbone}. Our RSFNet framework adopts the asymmetric encoder for learning RGB features and thermal features. In particular, we employ three encoder groups, \ie, RSFNet50-18, RSFNet161-12, and RSFNet101-34, where ResNet50 \cite{he-cvpr2016-resnet}, DenseNet161 \cite{huang-cvpr2017-densenet}, and ResNet101 \cite{he-cvpr2016-resnet} are the RGB encoders, while ResNet18 \cite{he-cvpr2016-resnet}, DenseNet121 \cite{huang-cvpr2017-densenet}, and ResNet34 \cite{he-cvpr2016-resnet} are the thermal encoders. 

\textbf{Training Phase}. All the encoders initially load the weights of the models pre-trained on the ImageNet \cite{deng-cvpr2009-imagenet} dataset, and the weights of other layers use the Kaiming initialization \cite{he-iccv2015-delving} parameters by PyTorch default. To adapt the pre-trained model weights for the thermal branch, we use the repeated thermal images along the channel dimension for the first convolutional layer. The initial learning rate $\ell_r$ is set to 0.01 with the polynomial learning rate decay, where $\ell_r$ is multiplied by $(1 - \frac{iter}{iter_{max}})^{power}$ with $power=0.9$. We adopt the stochastic gradient descent algorithm for model optimization, while the momentum and weight decay are set to 0.9 and 0.0005, respectively. Other settings are shown in Table~\ref{table:setting}, where the batch size and the epochs decide the $iter_{max}$. The kernel size $K$ is set to 5 for RSF module. In addition, we used some data augmentation skills, including random horizontal flipping, random cropping, and random rotation in $[-10^\circ, 10^\circ]$. The pixel values of all images are normalized to $[0, 1]$. 

\textbf{Inference Phase}. We simply feed the normalized test image to the already trained model, which yields the predicted label $\hat{\mathbf{Y}}$ as the segmentation results.

% -------------------------  Settings --------------------------
\begin{table}[!t]
	\centering
	\caption{Experimental settings.}
	\label{table:setting}
	\setlength{\tabcolsep}{0.9mm}{ 
	\begin{tabular}{lccccc}
	\toprule[0.75pt]
	\multirow{2}{*}{Dataset} & \multirow{2}{*}{Epoch} &\multicolumn{3}{c}{Batch Size} & \multirow{2}{*}{Crop Size} \\ 
	\cmidrule[0.5pt]{3-5}
	& & \scriptsize{RSFNet50-18} & \scriptsize{RSFNet161-121} & \scriptsize{RSFNet101-34} & \\
	\midrule[0.5pt]
	MFNet  \cite{ha-iros2017-mfnet}          & 300 & 8  & 4   & 4 & 256$\times$512 \\
	PST900 \cite{shivakumar-icra2020-pst900} & 100 & 4  & 2   & 2 & 360$\times$640 \\
	\toprule[0.75pt]
	\end{tabular}
}
\end{table}
	
\subsection{Compared Methods}
In total, we have compared three groups of semantic segmentation methods. 

\textit{RGB methods}. FRRN (Full-Resolution Residual Network) \cite{pohlen-cvpr2017-frrn}, PSPNet (Pyramid Scene Parsing Network) \cite{zhao-cvpr2017-pspnet}, BiseNetV1 (Bilateral Segmentation Network) \cite{yu-eccv2018-bisenet}, DFN (Discriminative Feature Network) \cite{yu-cvpr2018-dfn}, HRNet (High Resolution Network) \cite{sun-cvpr2019-hrnet}, BiseNetV2 \cite{yu-ijcv2021-bisenetv2}, and CAANet (Class Activation Attention Network) \cite{liu-cvpr2022-caanet}. These methods are also modified to adapt for the RGB-T images by increasing the input channel from 3 to 4 in the first convolutional layer. Note that HRNet uses itself, BiseNetV1 uses the ResNet18, and BiseNetV2 uses the convolutional network with 37 layers as the backbone; the remaining methods adopt the ResNet101 as the backbone.

\textit{RGB-T methods with indistinguishable fusion}. MFNet (Multispectral Fusion Network) \cite{ha-iros2017-mfnet}, RTFNet (RGB-Thermal Fusion Network) \cite{sun-ral2019-rtfnet}, PSTNet (Penn Subterranean Thermal Network) \cite{shivakumar-icra2020-pst900}, FuseSeg \cite{sun-tase2021-fuseseg}, and MFFENet (Multiscale Feature Fusion and Enhancement Network) \cite{huang-cvpr2017-densenet} (DenseNet121-121). 

\textit{RGB-T methods with distinguishable fusion}. ABMDRNet (Adaptive-weighted Bi-directional Modality Difference Reduction Network) \cite{zhang-cvpr2021-abmdrnet} (ResNet50-50), EGFNet (Edge-aware Guidance Fusion Network) \cite{zhou-aaai2022-egfnet}, and MTANet (MultiTask-Aware Network) \cite{zhou-tiv2022-mtanet}. 

All the above RGB-T methods adopt the symmetric encoders, which neglects the different ability of the RGB features and the thermal features that modeling the pixel-level relations under varying lighting conditions. By contrast, we use the asymmetric encoders at lower computational costs. Note that the backbone of the other methods can be found in Table~\ref{table:speed_modelPara}.

% --------------- Results on MFNet test set --------------
\begin{table*}[!t]
	\centering
	\caption{Comparison with state-of-the-art on MFNet test set. ``3c'' and ``4c'' denotes the RGB model and the RGB-T model, respectively; $^\ast$ denotes the reproduction results.}
	\label{table:result_mfnetdata}
	%\Huge
	\Large
	\resizebox{\textwidth}{!}{
	\begin{tabular}{l r c c c c c c c c c c c c c c c c c c c c c c c}
			\toprule[0.75pt]
			\multirow{2}{*}{Methods} & \multirow{2}{*}{Venue} & \multicolumn{2}{c}{Unlabeled} & \multicolumn{2}{c}{Car} & \multicolumn{2}{c}{Person} & \multicolumn{2}{c}{Bike}  & \multicolumn{2}{c}{Curve}  & \multicolumn{2}{c}{Car Stop}  & \multicolumn{2}{c}{Guardrail}  & \multicolumn{2}{c}{Color Cone}  & \multicolumn{2}{c}{Bump}  & \multirow{2}{*}{mAcc$\uparrow$}  & \multirow{2}{*}{mIoU$\uparrow$} \\
			\cmidrule[0.5pt]{3-20}  %\cline{3-20}
			%& & Acc & IoU & Acc & IoU & Acc & IoU & Acc & IoU & Acc & IoU & Acc & IoU & Acc & IoU & Acc & IoU & Acc & IoU \\
			& & {Acc} & {IoU} & {Acc} & {IoU} & {Acc} & {IoU} & {Acc} & {IoU} & {Acc} & {IoU} & {Acc} & {IoU} & {Acc} & {IoU} & {Acc} & {IoU} & {Acc} & {IoU} \\
			\midrule[0.5pt]
			\rowcolor{maroon}FRRN (3c)\cite{pohlen-cvpr2017-frrn} & CVPR'17 &  99.3 & 96.7 & 80.0 & 71.2 & 53.0 & 46.1 & 65.1 & 53.0 & 34.0 & 27.1 & 21.6 & 19.1 & 0.0 & 0.0 & 34.7 & 32.5 & 36.2 & 30.5 & 47.1 & 41.8 \\
			FRRN (4c) & CVPR'17 &  \textbf{99.5} & 96.9  & 81.9 & 74.7 & 66.2 & 60.8 & 62.8 & 50.3 & 41.2 & 35.0 & 12.5 & 11.5 & 0.0 & 0.0 & 37.2 & 34.0 & 35.2 & 34.6 & 48.5 & 44.2 \\
			\rowcolor{maroon}PSPNet$^{\ast}$ (3c)\cite{zhao-cvpr2017-pspnet} & CVPR'17 & 98.7 & 97.2 & 94.2 & 82.6 & 67.6 & 54.1 & 70.7 & 55.8 & 28.1 & 22.1 & 26.9 & 23.3 & 3.4 & 1.1 & 37.3 & 33.1 & 40.6 & 33.5 & 51.9 & 44.7 \\
			PSPNet$^{\ast}$ (4c) & CVPR'17 & 98.9 & 97.4 & 94.0 & 84.7 & 69.4 & 57.5 & 71.6 & 56.2 & 27.7 & 22.9 & 26.7 & 23.3 & 20.8 & 4.4 & 36.3 & 32.5 & 40.7 & 37.4 & 54.0 & 46.3 \\
			\rowcolor{maroon}BiseNetV1$^{\ast}$ (3c)\cite{yu-eccv2018-bisenet} & ECCV'18 & 99.2 & 97.6 & 90.4 & 82.5 & 70.2 & 58.5 & 65.6 & 57.1 & 38.2 & 29.6 & 27.8 & 23.1 & 2.6 & 1.1 & 43.9 & 39.8 & 46.6 & 43.5 & 53.8 & 48.1 \\
			BiseNetV1$^{\ast}$ (4c) & ECCV'18 & 99.3 & 97.9 & 90.1 & 83.4 & 77.3 & 69.3 & 66.0 & 57.4 & 49.9 & 40.3 & 28.7 & 23.9 & 7.6 & 2.8 & 38.5 & 36.9 & 44.9 & 44.0 & 55.8 & 50.7 \\
			\rowcolor{maroon}DFN (3c)\cite{yu-cvpr2018-dfn} & CVPR'18 & 98.6 & 97.7 & 90.7 & 81.4 & 67.7 & 52.8 & 71.5 & 57.5 & 49.2 & 34.9 & 35.1 & 23.8 & 4.1 & 0.9 & 44.2 & 31.0 & 54.6 & 47.5 & 57.3 & 47.5 \\
			DFN (4c) & CVPR'18 & 98.6 & 97.7 & 90.0 & 84.4 & 73.2 & 65.0 & 75.5 & 60.9 & 54.0 & 40.4 & 38.9 & 25.7 & 10.2 & 4.0 & 48.3 & 42.5 & 55.8 & 47.4 & 60.5 & 52.0 \\
			\rowcolor{maroon}HRNet (3c)\cite{sun-cvpr2019-hrnet} & CVPR'19 & 99.2 & 97.3 & 92.2 & 86.6 & 73.1 & 59.8 & 74.9 & 61.3 & 47.0 & 33.2 & 38.3 & 28.7 & 7.3 & 1.4 & 54.6 & 47.2 & 61.5 & 46.2 & 60.9 & 51.3 \\
			HRNet (4c) & CVPR'19 & 98.6 & \underline{98.2} & 92.8 & 87.6 & 79.3 & 71.0 & 78.3 & \underline{63.4} & 59.8 & 42.5 & 25.7 & 19.1 & 18.8 & 2.7 & 56.5 & 49.8 & 63.5 & 44.5 & 63.7 & 53.2 \\
			\rowcolor{maroon}BiseNetV2$^{\ast}$ (3c)\cite{yu-ijcv2021-bisenetv2} & IJCV'21 & 98.3 & 96.0 & 82.8 & 63.1 & 50.1 & 41.3 & 49.2 & 41.4 & 13.9 & 11.4 & 9.8 & 7.9 & 0.0 & 0.0 & 22.2 & 20.5 & 25.6 & 20.0 & 39.1 & 33.5 \\
			BiseNetV2$^{\ast}$ (4c) & IJCV'21 & 98.4 & 96.3 & 82.5 & 67.4 & 69.1 & 57.7 & 46.3 & 36.3 & 24.0 & 21.1 & 7.5 & 5.8 & 0.0 & 0.0 & 17.8 & 16.6 & 22.6 & 20.7 & 40.9 & 35.8 \\
			\rowcolor{maroon}CAANet$^{\ast}$ (3c)\cite{liu-cvpr2022-caanet} & CVPR'22 & 99.2 & 97.8 & 91.3 & 85.7 & 72.8 & 61.6 & 68.2 & 59.2 & 46.6 & 34.4 & 29.8 & 25.9 & 9.1 & 4.0 & 51.6 & 46.7 & 50.1 & 47.3 & 57.6 & 51.4 \\
			CAANet$^{\ast}$ (4c) & CVPR'22 & 99.3 & 98.0 & 91.4 & 87.3 & 80.8 & 71.2 & 69.3 & 59.1 & 51.4 & 40.1 & 27.0 & 24.1 & 2.3 & 0.8 &  52.8 & 49.4 & 61.4 & 49.2 & 59.5 & 53.2 \\
				\midrule[0.5pt]
			\rowcolor{maroon}MFNet\cite{ha-iros2017-mfnet} & IROS'17 & 98.7 & 96.9 &  77.2 & 65.9 & 67.0 & 58.9 & 53.9 & 42.9 & 36.2 & 29.9 & 12.5 & 9.9 & 0.1 & 0.0 & 30.3 & 25.2 & 30.0 & 27.7 & 45.1 & 39.7 \\
			RTFNet\cite{sun-ral2019-rtfnet} & RAL'19 & \underline{99.4} & \textbf{98.5} & 93.0 & 87.4 & 79.3 & 70.3 & 76.8 & 62.7 & 60.7 & \underline{45.3} & 38.5 & 29.8 & 0.0 & 0.0 & 45.5 & 29.1 & \underline{74.7} & \textbf{55.7} & 63.1 & 53.2 \\
			\rowcolor{maroon}PSTNet$^{\ast}$\cite{shivakumar-icra2020-pst900} & ICRA'20 & 98.8 & 97.4 & 89.4 & 77.0 & 74.8 & 67.2 & 65.8 & 50.8 & 42.5 & 35.2 & 34.6 & 25.2 & 0.0 & 0.0 & 32.7 & 29.9 & 61.1 & \underline{55.0} & 55.5 & 48.6 \\
			FuseSeg\cite{sun-tase2021-fuseseg} & TASE'21 & 99.0 & 97.6 & 93.1 &  \underline{87.9} & 81.4 & 71.7 & 78.5 & \textbf{64.6} &  68.4 &  44.8 & 29.1 & 22.7 & \textbf{63.7} & 6.4 & 55.8 & 46.9 & 66.4 & 47.9 & 70.6 & 54.5 \\
			\rowcolor{maroon}ABMDRNet\cite{zhang-cvpr2021-abmdrnet} & CVPR'21 & 98.6 & 97.8 & 94.3 & 84.8 & 90.0 & 69.6 & 75.7 & 60.3 & 64.0 & 45.1 & 44.1 & 33.1 & 31.0 & 5.1 & 61.7 & 47.4 & 66.2 & 50.0 & 69.5 & 54.8 \\
			MFFENet\cite{zhou-tmm2021-mffenet} & TMM'22 & 99.3 & 97.8 & 91.4 & 87.1 & 82.6 & \textbf{74.4} & 76.7 & 61.3 & 58.7 & \textbf{45.6} & 44.9 & 30.6 & 60.0 & 5.2 & 64.4 & \textbf{57.0} & 72.7 & 40.5 & 72.3 & 55.5 \\
			\rowcolor{maroon}EGFNet\cite{zhou-aaai2022-egfnet} & AAAI'22 & 98.7 & 98.0 & \underline{95.8} & 87.6 & 89.0 & 69.8 & 80.6 & 58.8 & 71.5 & 42.8 & \underline{48.7} & \underline{33.8} & 33.6 & 7.0 & \underline{65.3} & 8.3 & 71.1 & 47.1 & 72.7 & 54.8 \\
			MTANet\cite{zhou-tiv2022-mtanet} & TIV'22 & 98.4 & 98.0 & \underline{95.8} & \textbf{88.1} & \underline{90.9} & 71.5 & 80.3 & 60.7 & \underline{75.3} & 40.9 & \textbf{62.8} & \textbf{38.9} & 38.7 & \textbf{13.7} & 63.8 & 45.9 & 70.8 & 47.2 & \underline{75.2} & \underline{56.1} \\
			\midrule[0.5pt]
			
			\rowcolor{maroon}Ours (ResNet50-18) & & 96.4 & 97.7 & 93.4 & 83.5 & \textbf{91.8} & 72.7 & \underline{83.7} & 61.9 & 75.1 & 40.7 & 39.7 & 31.7 & 59.1 & \underline{11.2} & 59.1 & 50.9 & 60.5 & 45.5 & 73.2 & 55.1 \\				
			Ours ({\large DenseNet161-121}) & & 98.1 & 97.9 & 94.6 & 86.3 & 89.7 & \underline{72.9} & \textbf{88.2} & 61.7 & 60.3 & 43.7 & 27.2 & 30.8 & 56.7 & 6.5 & \textbf{68.4} & 52.4 & 81.3 & 47.3 & 73.8 & 55.5 \\
			\rowcolor{maroon}Ours (ResNet101-34) & & 97.3 & 97.9 & \textbf{95.9} & 85.8 & 90.8 & 72.0 & 81.4 & 60.3 & \textbf{82.1} & 43.3 & 30.9 & 32.3 & \underline{63.0} & 8.8 & 65.0 & \underline{52.9} & \textbf{76.2} & 52.7 & \textbf{75.9} & \textbf{56.2} \\
			\toprule[0.75pt]
	\end{tabular}
}
\end{table*}

% --------------- Inference speed and model parameters on MFNet test set --------------
\begin{table}[!t]
	\centering
	\caption{Comparison of inference speed and model parameters on MFNet test set. ``5Conv'' means 5 convolution layers; ``$^{\ast}$'' denotes the reproduction results; `` -'' means it is unavailable.}
	\label{table:speed_modelPara}
	%\resizebox{\textwidth}{!} 	{  % resize the entire table 
	\setlength{\tabcolsep}{0.8mm}{  % control the table size on a whole
	\begin{tabular}{l  l r r r r}
		\toprule[0.75pt]
		\multirow{2}{*}{Methods}  & \multirow{2}{*}{Backbone} & \multirow{2}{*}{$\#$Params$\downarrow$} & \multirow{2}{*}{FLOPs$\downarrow$} & \multicolumn{2}{c}{FPS$\uparrow$} \\
		\cmidrule[0.5pt]{5-6} % 
		%\cline{8-9}
		&   & (M) & (G) & 2080Ti & TITAN Xp \\
		\midrule[0.5pt]
		\rowcolor{maroon}MFNet$^{\ast}$\cite{ha-iros2017-mfnet}  & 5Conv-5Conv & \textbf{0.74} & \textbf{8.42} & \textbf{205.47} & \textbf{137.39} \\
		RTFNet$^{\ast}$\cite{sun-ral2019-rtfnet} & ResNet152-152  & 254.51 & 336.69 & 33.86 & 14.63  \\
		\rowcolor{maroon}PSTNet$^{\ast}$\cite{shivakumar-icra2020-pst900} & ResNet18 & \underline{31.04} & 163.64 & \underline{102.82} & \underline{64.76} \\
		FuseSeg\cite{sun-tase2021-fuseseg}  & DenseNet161-161 & 141.52 & 193.40 & - & 16.88  \\
		\rowcolor{maroon}EGFNet$^{\ast}$\cite{zhou-aaai2022-egfnet}  & ResNet152-152& 62.78 & 201.09 & 18.36 & 8.12 \\
		MTANet\cite{zhou-tiv2022-mtanet}  & ResNet152-152  & 121.58 & 264.96 & - & 8.40 \\
		\midrule[0.5pt]
		Ours   & ResNet50-18 & 44.53 & \underline{64.69} & 81.36 & 40.47 \\
		\rowcolor{maroon}		Ours  & DenseNet161-121  & 44.10 & 95.02 & 34.15 & 14.10 \\
		Ours  & ResNet101-34  & 73.63 & 98.82 & 57.31 & 28.40 \\
		\toprule[0.75pt]
	\end{tabular}
	}
\end{table}

\begin{table}[!t]
	\centering
	\caption{Performance comparison on the daytime and nighttime images of MFNet test set.}
	\label{table:day-night-mfnetdata}
	\setlength{\tabcolsep}{1mm}{  % control the table size on a whole
	\begin{tabular}{l r c c c c}
		\toprule[0.75pt]
		\multirow{2}{*}{Methods} & \multirow{2}{*}{Venue} &  \multicolumn{2}{c}{Daytime} & \multicolumn{2}{c}{Nighttime} \\
		\cmidrule[0.5pt]{3-6} % \cline{3-6}
		& & mAcc$\uparrow$ & mIoU$\uparrow$ & mAcc$\uparrow$ & mIoU$\uparrow$  \\
		\midrule[0.5pt]
		\rowcolor{maroon}FRRN (3c)\cite{pohlen-cvpr2017-frrn} & CVPR'17 & 45.1 & 40.0 & 41.6 & 37.3 \\
		FRRN (4c) & CVPR'17 & 42.4 & 38.0 & 46.2 & 42.3 \\
		\rowcolor{maroon}PSPNet$^{\ast}$ (3c)\cite{zhao-cvpr2017-pspnet} & CVPR'17 & 49.3 & 41.9 & 47.8 & 41.5 \\
		PSPNet$^{\ast}$ (4c) & CVPR'17 & 47.2 & 41.0 & 50.5 & 44.0 \\
		\rowcolor{maroon}BiseNetV1$^{\ast}$ (3c)\cite{yu-eccv2018-bisenet} & ECCV'18 & 51.3 & 44.4 & 48.8 & 43.8 \\
		BiseNetV1$^{\ast}$ (4c) & ECCV'18 & 50.6 & 45.0 & 52.5 & 47.8 \\
		\rowcolor{maroon}DFN (3c)\cite{yu-cvpr2018-dfn} & CVPR'18 & 53.7 & 42.2 & 52.4 & 44.6 \\
		DFN (4c) & CVPR'18 & 53.4 & 43.9 & 57.4 & 51.8 \\
		\rowcolor{maroon}HRNet (3c)\cite{sun-cvpr2019-hrnet} & CVPR'19 & 59.7 & 47.2 & 55.7 & 49.1 \\
		HRNet (4c) & CVPR'19 & 50.0 & 41.4 & 50.2 & 44.9 \\
		\rowcolor{maroon}BiseNetV2$^{\ast}$ (3c)\cite{yu-ijcv2021-bisenetv2} & IJCV'21 & 37.2 & 32.0 & 34.8 & 29.7 \\
		BiseNetV2$^{\ast}$ (4c) & IJCV'21 & 35.8 & 30.8 & 39.7 & 34.5 \\
		\rowcolor{maroon}CAANet$^{\ast}$ (3c)\cite{liu-cvpr2022-caanet} & CVPR'22 & 52.9 & 46.2 & 53.1 & 47.4 \\
		CAANet$^{\ast}$ (4c) & CVPR'22 & 52.3 & 46.5 & 55.4 & 50.9 \\ 
		\midrule[0.5pt]
		\rowcolor{maroon}MFNet\cite{ha-iros2017-mfnet} & IROS'17 & 42.6 & 36.1 & 41.4 & 36.8 \\
		RTFNet\cite{sun-ral2019-rtfnet} & RAL'19 & 60.0 & 45.8 & 60.7 & 54.8 \\
		\rowcolor{maroon}PSTNet$^{\ast}$\cite{shivakumar-icra2020-pst900} & ICRA'20 & 45.3 & 39.6 & 55.8 & 48.2 \\
		FuseSeg\cite{sun-tase2021-fuseseg} & TASE'21 & 62.1 & 47.8 & 67.3 & 54.6 \\
		\rowcolor{maroon}MFFENet\cite{zhou-tmm2021-mffenet} & TMM'22 & 70.5 & \underline{47.9} & 70.0 & 56.7 \\
		EGFNet\cite{zhou-aaai2022-egfnet} & AAAI'22 & \underline{74.4} & 47.3 & 67.0 & 55.0 \\
		\rowcolor{maroon}MTANet\cite{zhou-tiv2022-mtanet} & TIV'22 & \textbf{77.7} & 47.5 & 72.8 & \textbf{58.0} \\
		\midrule[0.5pt]
		Ours (ResNet50-18) & & 69.1 & 47.6 & 71.2 & 55.8\\
		\rowcolor{maroon}Ours (DenseNet161-121) & & 71.8 & 47.7 & \underline{73.0} & 56.1 \\
		Ours (ResNet101-34) & & 72.9 & \textbf{49.1} & \textbf{73.8} & \underline{57.2} \\
		\toprule[0.75pt]
	\end{tabular}
}
\end{table}

% -------------- Result on PST900 test set -------------
\begin{table}[!t]
	\centering
	\caption{Performance comparison on PST900 test set.}
	\label{table:result_pst900}
	%\resizebox{0.5\textwidth}{!}{}
	\setlength{\tabcolsep}{0.7mm}{
	\begin{tabular}{l r c c c c c c}
			\toprule[0.75pt]
			Methods & Venue & \scriptsize{Backgd} & \scriptsize{FireExt.} & \scriptsize{Backpk} & \scriptsize{HandDri.}  & \scriptsize{Survi.}  & mIoU$\uparrow$ \\
		
			\midrule[0.5pt]
			\rowcolor{maroon}
			ERFNet (3c)\cite{romera-tits2017-erfnet} & \scriptsize{TITS'17} & 98.7 & 61.2 & 65.3 & 42.4 & 41.7 & 61.9 \\
			ERFNet (4c) & \scriptsize{TITS'17} & 98.7 & 58.8 & 68.1 & 52.8 & 34.3 & 62.6 \\
			\rowcolor{maroon}
			PSPNet$^{\ast}$(3c)\cite{zhao-cvpr2017-pspnet} & \scriptsize{CVPR'17} & 99.1 & 54.0 & 79.8 & 54.9 & 56.1 & 68.7 \\
			PSPNet$^{\ast}$(4c) & \scriptsize{CVPR'17} & 99.1 & 62.5 & 79.8 & 53.4 & 58.1 & 70.5 \\
			\rowcolor{maroon}
			\scriptsize{BiseNetV1$^{\ast}$(3c)\cite{yu-eccv2018-bisenet}} & \scriptsize{ECCV'18} & 99.0 & 68.0 & 72.0 & 47.0 & 5.7 & 68.4 \\
			\scriptsize{BiseNetV1$^{\ast}$(4c)} & \scriptsize{ECCV'18} & 98.8 & 73.0 & 64.8 & 52.0 & 57.1 & 69.1 \\
			\rowcolor{maroon}
			\scriptsize{BiseNetV2$^{\ast}$(3c)\cite{yu-ijcv2021-bisenetv2}} & \scriptsize{IJCV'21} & 98.4 & 47.9 & 54.2 & 21.3 & 14.6 & 47.3 \\
			\scriptsize{BiseNetV2$^{\ast}$(4c)} & \scriptsize{IJCV'21} & 98.4 & 47.5 & 59.8 & 22.6 & 16.0 & 48.9 \\
			\rowcolor{maroon}
			CAANet$^{\ast}$(3c)\cite{liu-cvpr2022-caanet} & \scriptsize{CVPR'22} &  98.7 & 62.0 & 66.2 & 66.7 & 59.3 & 70.6 \\
			CAANet$^{\ast}$(4c) & \scriptsize{CVPR'22} & 98.0 & 71.7 & 47.8 & 71.4 & 68.9 & 71.6 \\
				\midrule[0.5pt]
			\rowcolor{maroon}
			MFNet\cite{ha-iros2017-mfnet} & \scriptsize{IROS'17} & 98.6 & 60.4 & 64.3 & 41.1 & 20.7 & 57.0 \\
			RTFNet$^{\ast}$\cite{sun-ral2019-rtfnet} & \scriptsize{RAL'19} & \underline{99.3} & \underline{74.1} & 79.7 & 58.8 & 61.1 & 74.6 \\
			\rowcolor{maroon}
			PSTNet\cite{shivakumar-icra2020-pst900} & \scriptsize{ICRA'20} & 98.9 & 70.1 & 69.2 & 53.6 & 50.0 & 68.4  \\
			MFFENet\cite{zhou-tmm2021-mffenet} & \scriptsize{TMM'22} & \underline{99.3} & 79.8 & 76.6 & 66.8 & 63.0 & 77.1 \\
			\rowcolor{maroon}
			EGFNet\cite{zhou-aaai2022-egfnet} & \scriptsize{AAAI'22}  & 99.2 & 71.3 & \underline{83.1} & 64.7 & \textbf{74.3} & 78.5
			\\
			MTANet\cite{zhou-tiv2022-mtanet} & \scriptsize{TIV'22} & \underline{99.3} & 65.0 & 87.5 & 62.1 & 79.1 & 78.6 \\
			\midrule[0.5pt]
			\rowcolor{maroon}Ours (\scriptsize{ResNet50-18}) & &  \underline{99.3} & 68.7 & 81.9 & 69.7 & 73.1 & 78.5 \\
			Ours (\tiny{DenseNet161-121}) & & \underline{99.3} & 72.0 & 80.4 & \textbf{74.5} & \underline{70.8} & \underline{79.4} \\
			\rowcolor{maroon}Ours (\scriptsize{ResNet101-34}) & & \textbf{99.4} & \textbf{75.4} & \textbf{84.9} & \underline{72.9} & 70.1 & \textbf{80.5} \\
			\toprule[0.75pt]
	\end{tabular}
}
\end{table}

% -------------------------------
\subsection{Quantitative Results}
We show the results on MFNet dataset and PST900 dataset, respectively. The records of the compared methods are taken from the original papers or reproduced when specified by ``$\ast$''. The best records are highlighted in boldface and the second-best ones are underlined.

% -------------  Figure of inference speed, model size, and mIoU -----------
\begin{figure}[!t]
	\centering
	\includegraphics[width=0.3\textwidth]{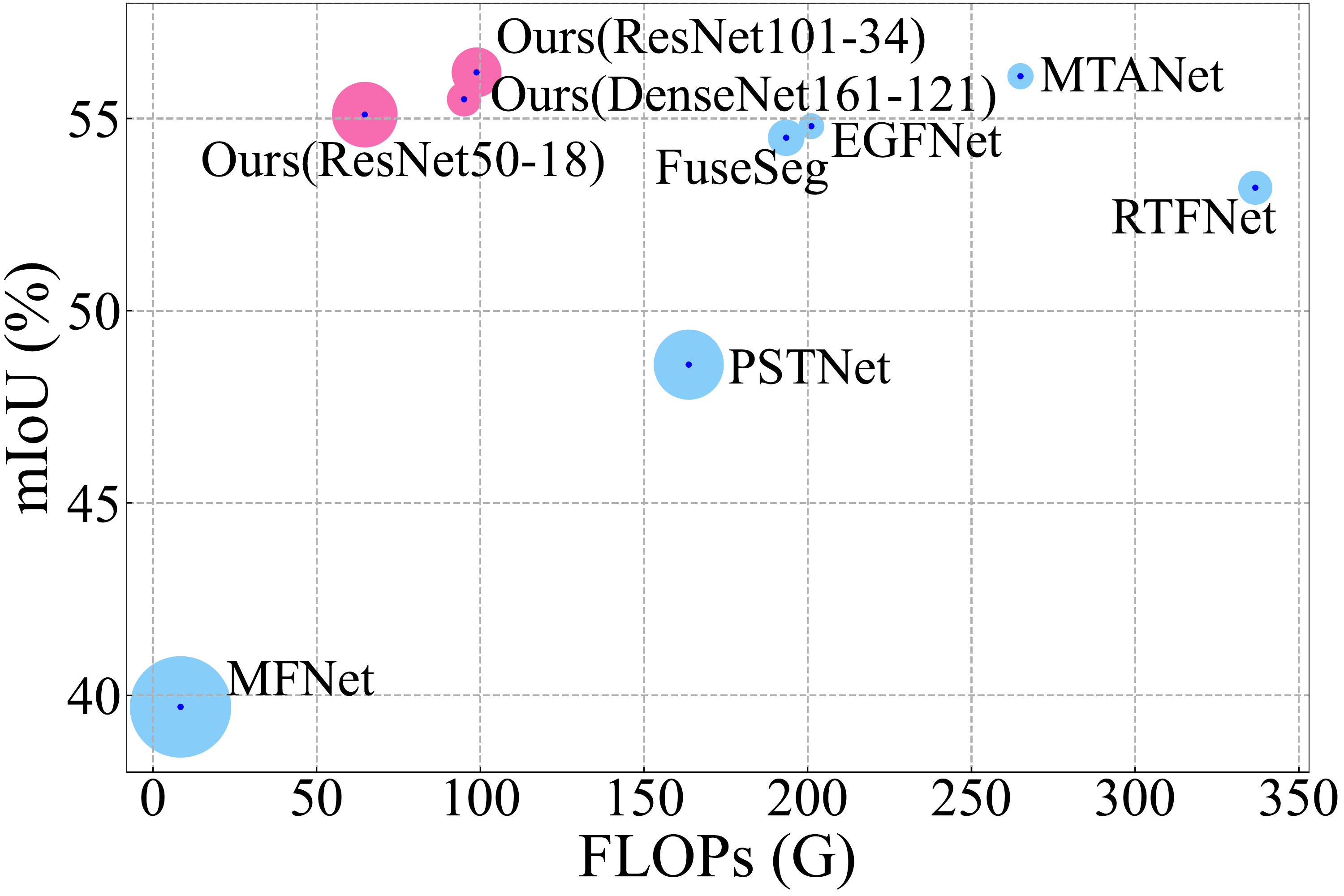}  %, height=0.25\textwidth
	\caption{Inference speed (FPS), model parameters (FLOPs), and segmentation performance (mIoU) on MFNet test set. The inference speed was evaluated using one TITAN Xp graphics card, the larger the circle, the faster the speed. }
	\label{fig:inference_speed}
\end{figure}

\textbf{MFNet Dataset}. The comparison results are recorded in Table~\ref{table:result_mfnetdata}. Meanwhile, the comparison of the inference speed and the model parameter are tabulated in Table~\ref{table:speed_modelPara}, which is also illustrated in Fig.~\ref{fig:inference_speed}. Note that MFFENet \cite{huang-cvpr2017-densenet} and ABMDRNet \cite{zhang-cvpr2021-abmdrnet} are not reported because their models are unavailable. In addition, we show the results of the daytime and nighttime scenarios in Table~\ref{table:day-night-mfnetdata}. From these tables and figures, we have the following observations.
\begin{itemize}
	\item Our RSFNet method with ResNet101-34 as the backbone consistently outperforms all the compared ones, and also strikes a good balance between the inference speed and the performance. It achieves the real-time segmentation at nearly 60 fps and 30 fps on 2080Ti and TITAN Xp graphics card, respectively. This demonstrates that the proposed asymmetric encoder can save lost of computational costs, and the Residual Spatial Fusion module effectively fuses the RGB and the thermal features, which are learned under the pseudo-label guidance.
	
	\item Our method is slightly better than MTANet \cite{zhou-tiv2022-mtanet} (Table~\ref{table:result_mfnetdata}), but the model parameters and the FLOPs are only its 60\% and 37\% (Table~\ref{table:speed_modelPara}), respectively. Moreover, our inference speed is nearly four times faster.
	
	\item Those methods with three channels perform poorer than that with four channels in Table~\ref{table:result_mfnetdata}, \eg, DFN-4c \cite{yu-cvpr2018-dfn} has a 4.5\% gain compared to DFN-3c. This verifies the fact that the thermal modality does help improve the segmentation performance in most scenarios.
	
	\item MFNet \cite{ha-iros2017-mfnet} requires the least model parameters and achieves the largest inference speed in Fig.~\ref{fig:inference_speed}, but it has very poor segmentation performance, about 30\% lower than ours in terms of mAcc. This is because it adopts only 5 convolution layers to encode the dual-modality features and does a simple concatenation fusion.
	
	\item The performances in daytime are consistently better than that in nighttime for all methods in Table~\ref{table:day-night-mfnetdata}. This suggests that good lighting conditions are important for segmentation. The three-channel models perform better than four-channel ones in daytime, while they perform worse than four-channel models at night, which indicates that RGB images dominate the segmentation in daytime and thermal images contribute more at night. Moreover, our RSFNet outperforms the others in terms of mIoU on daytime images and in terms of mAcc on nighttime images, which validates that the proposed fusion network strategy works well for all-day lighting conditions.
		
\end{itemize}

\textbf{PST900 Dataset}. The results on PST900 dataset is recorded in Table~\ref{table:result_pst900}, where we have the similar observations to MFNet dataset. Our RSFNet method consistently performs better in terms of mIoU across almost all classes, \eg, ours-ResNet101-34 achieves a gain of 1.9\% over MTANet \cite{zhou-tiv2022-mtanet}, whose model size is much larger than ours. This also shows the advantages of our method in the challenging underground environments.

% -------------------------------
\subsection{Ablation Studies}
Except for the special description, we use ResNet50-18 as the backbone of RSFNet without structural re-parameterization to do the ablation studies on the test set, and the parameters or settings are the same to that of training.

\textbf{RSF module}. We investigate the influences of the pseudo-label supervision and the structural re-parameterization, whose results are shown in Table~\ref{table:abl_rsfnet}. When applying the structural re-parameterization ($row2$), the performance is improved by 1.7\% and 1.8\% in terms of mAcc on MFNet dataset and PST900 dataset, respectively, without additional costs. When adopting the pseudo-label supervision ($row3$), the performance is boosted by 1.0\% in terms of mAcc on the two datasets. And the best results are achieved by using both of them ($row4$). This verifies the advantage of the structural re-parameterization in RSF module and the pseudo-label supervision is beneficial for capturing the saliency-aware cross-modality features. 

% ---------- Albation of RSF module ----------
\begin{table}[!t]
	\centering
	\caption{Ablation of the Residual Spatial Fusion module. Stru.: Structure Re-parameterization; Pse.: pseudo label for supervision. FLOPs(G): 64.69 (MFNet) and 194.56 (PST900).}
	\label{table:abl_rsfnet}
	\begin{tabular}{c c c c  c c  c c}
		\toprule[0.75pt]
		\multirow{2}{*}{Stru.} & \multirow{2}{*}{Pse.} & \multicolumn{2}{c}{MFNet Dataset} & &
		\multicolumn{2}{c}{PST900 Dataset} & \multirow{2}{*}{$\#$Params} \\
		\cmidrule[0.5pt]{3-4} \cmidrule[0.5pt]{6-7}
		& & mAcc$\uparrow$ & mIoU$\uparrow$  & & mAcc$\uparrow$ & mIoU$\uparrow$  &(M)$\downarrow$  \\
		
		\midrule[0.5pt]
		\rowcolor{maroon}& & 70.4 & 53.3  & & 82.2 & 77.0  & \textbf{44.20} \\
		
		\checkmark & & \underline{72.1} & \underline{54.3} && \underline{84.0} & \underline{77.6}  & \textbf{44.20} \\
		
		\rowcolor{maroon}& \checkmark & 71.4 & 54.0  && 83.2 & 77.5 & \underline{44.53} \\
		
		\checkmark & \checkmark & \textbf{73.2} & \textbf{55.1}  && \textbf{85.2} & \textbf{78.5}  & \underline{44.53} \\			
		\toprule[0.75pt]
	\end{tabular}
	%}
\end{table}

\textbf{Fusion Scheme}. We tried three several fusion strategies for fusing the RGB and the thermal features, including two indistinguishable schemes, \ie, element-wise summation and simple concatenation, RSF without pseudo-label supervision, RSF, and Gated Multi-modality Attention (GMA) \cite{chen-tip2020-dpanet}. The results are shown in Table~\ref{table:abl_fusion}, from which we see that RSF outperforms the element-wise summation by 9.3\% and 1.4\% in terms of mAcc and mIoU on MFNet dataset. When the pseudo-label supervision is considered, the performance is boosted by 1.0\% in terms of mAcc on the two datasets. This demonstrates the superiority of the RSF fusion scheme and the pseudo-label guidance.

% ---------- Different fusion schemes using ResNet50-18 -------------
\begin{table}[!t]
	\centering
	\caption{Different fusion schemes. Pse.: pseudo-label for supervision; Sum.: element-wise summation;  Conc.: concatenation; RSF: Residual Spatial Fusion; GMA: Gated Multi-modality Attention \cite{chen-tip2020-dpanet}.} %
	\label{table:abl_fusion}
	%\resizebox{0.5\textwidth}{!}{}
	\setlength{\tabcolsep}{0.7mm}{
		\begin{tabular}{l c c c c c c c c c}
			\toprule[0.75pt]
			\multirow{2}{*}{Fusion} & \multirow{2}{*}{Pse.} & \multicolumn{3}{c}{MFNet Dataset} & &
			\multicolumn{3}{c}{PST900 Dataset} & \multirow{2}{*}{$\#$Params} \\
			\cmidrule[0.5pt]{3-5} \cmidrule[0.5pt]{7-9}
			& & mAcc$\uparrow$ & mIoU$\uparrow$ & FLOPs(G)$\downarrow$ & & mAcc$\uparrow$ & mIoU$\uparrow$ & FLOPs(G)$\downarrow$ & (M)$\downarrow$  \\
			\midrule[0.5pt]
			\rowcolor{maroon}Sum. &  & 61.1 & 51.9 & \textbf{58.60} & & 77.0 & 73.6 & \textbf{176.27} & \textbf{43.11} \\
			Conc. & & 62.1 & 52.2 & \underline{59.11} & & 77.9 & 74.3 & \underline{177.81} & \underline{43.42} \\
			\rowcolor{maroon}RSF & & \underline{70.4} & \underline{53.3} & 64.69 & & \underline{82.2} & \underline{77.0} & 194.56 & 44.20 \\
			RSF & \checkmark & \textbf{71.4} & \textbf{54.0} & 64.69 & & \textbf{83.2} & \textbf{77.5} & 194.56 & 44.53 \\
			\rowcolor{maroon}GMA & $\checkmark$ & 68.6 & 52.8 & 73.80 & & 81.6 & 76.5 & 222.37 & 52.05 \\
			\toprule[0.75pt]
		\end{tabular}
	}
\end{table}

\textbf{Kernel Size $K$}. We vary the kernel size $K$ in the RSF module from 3 to 9 with an interval of 2, and the results are shown in Table~\ref{table:abl_kernlSize}. It can observed that the best size is 5, and the smaller one or the larger one will weaken the performance. This suggest that the local patterns captured spatially are prone to an intermediate kernel size of the convolution layer. 

% -------------- Kernel size K in RSF module -------------
\begin{table}[!t]
	\centering
	\caption{Ablation of the kernel size $K$ in RSF module. }
	\label{table:abl_kernlSize}
	%\resizebox{0.5\textwidth}{!}{}
	\setlength{\tabcolsep}{0.9mm}{
		\begin{tabular}{c c c c c c c c c}
			\toprule[0.75pt]
			\multirow{2}{*}{$K$} & \multicolumn{3}{c}{MFNet Dataset} & & \multicolumn{3}{c}{PST900 Dataset} & \multirow{2}{*}{$\#$Params} \\
			\cmidrule[0.5pt]{2-4} \cmidrule[0.5pt]{6-8}
			& mAcc$\uparrow$ & mIoU$\uparrow$ & FLOPs(G)$\downarrow$ & & mAcc$\uparrow$ & mIoU$\uparrow$ & FLOPs(G)$\downarrow$ & (M)$\downarrow$ \\	
			\midrule[0.5pt]
			\rowcolor{maroon}3 & 69.5 & 53.2 & \textbf{61.35} && 82.1 & 77.0 & \textbf{184.53} & \textbf{44.01} \\
			%\cline{2-6}
			5 & \textbf{71.4} & \textbf{54.0} & \underline{64.69} && \textbf{83.2} & \textbf{77.5} & \underline{194.56} & \underline{44.53} \\
			%\cline{2-6}
			\rowcolor{maroon}7 & \underline{71.0} & \underline{53.8} & 69.70 && \underline{82.8} & \underline{77.4} & 209.60 & 45.32 \\
			%\cline{2-6}
			9 & 70.5 & 53.7 & 76.39 && 82.7 & 77.3 & 229.66 & 46.37 \\
			\toprule[0.75pt]
		\end{tabular}
	}
\end{table}

\textbf{Reduced Dimension $\tilde{C}$}. This examines the impact of the reduced dimension of the feature $\mathbf{Z}_s$ in the cross-modality fusion. We vary the dimension from 32 to 256 with an equal ratio 2, and show the results in Table~\ref{table:abl_reduceDim_c}. It suggests that when $\tilde{C}$ is 64, our RSFNet achieves the best performance with relatively low FLOPs.

% ---------- Reduced number $\tilde{C}$  -------------
\begin{table}[!t]
	\centering
	\caption{Ablation of reduced dimension $\tilde{C}$ using RSFNet50-18. }
	\label{table:abl_reduceDim_c}
	\setlength{\tabcolsep}{0.72mm}{
		\begin{tabular}{c c c c c c c c c}
			\toprule[0.75pt]
			\multirow{2}{*}{$\tilde{C}$} & \multicolumn{3}{c}{MFNet Dataset} & & \multicolumn{3}{c}{PST900 Dataset} & \multirow{2}{*}{$\#$Params$\downarrow$} \\
			\cmidrule[0.5pt]{2-4} \cmidrule[0.5pt]{6-8}
			& mAcc$\uparrow$ & mIoU$\uparrow$ & FLOPs(G)$\downarrow$ & & mAcc$\uparrow$ & mIoU$\uparrow$ & FLOPs(G)$\downarrow$ & (M) \\	
			\midrule[0.5pt]
			\rowcolor{maroon}32 & 69.2 & 53.0 & \textbf{60.34} && 81.5 & 76.7 & \textbf{181.51} & \textbf{43.78} \\
			64 & \textbf{71.4} & \textbf{54.0} & \underline{64.69} && \textbf{83.2} & \textbf{77.5} & \underline{194.56} & \underline{44.53} \\
			\rowcolor{maroon}128 & \underline{71.2} & \underline{53.9} & 81.21 && \underline{83.0} & \underline{77.3} & 244.16 & 47.26 \\
			256 & 70.8 & 53.7 & 145.61 & &  82.7 & 77.0 & 437.39 & 57.63 \\
			\toprule[0.75pt]
		\end{tabular}
	}
\end{table}

\textbf{Regression Loss and $\lambda$}. We examine the influences of different forms of the regression loss and the regularization hyper-parameter $\lambda$ in the range $[0.1:0.1:0.5]$, whose results are respectively shown in Table~\ref{table:abl_regloss} and Table~\ref{table:abl_lambda}. From these records, we see that the Smooth L1-loss is the best choice for regressing the learned features to an estimated confidence score, which is supervised by the pseudo-label. Moreover, the hyper-parameter that controls the segmentation loss and the regression loss should be neither too large nor too small. 

% ------------ Different regression loss -------
\begin{table}[!t]
	\centering
	\caption{Different regression loss.}
	\label{table:abl_regloss}
	\setlength{\tabcolsep}{2.5mm}{
		\begin{tabular}{l c c c c c}
			\toprule[0.75pt]
			\multirow{2}{*}{Loss function} & \multicolumn{2}{c}{MFNet Dataset} & & \multicolumn{2}{c}{PST900 Dataset} \\
			\cmidrule[0.5pt]{2-3} \cmidrule[0.5pt]{5-6}
			& mAcc$\uparrow$ & mIoU$\uparrow$ & & mAcc$\uparrow$ & mIoU$\uparrow$ \\ 
			\midrule[0.5pt]
			\rowcolor{maroon}Smooth L1-loss & \textbf{71.4} & \textbf{54.0} && \textbf{83.2} & \textbf{77.5}  \\
			L1-loss & 70.4 & 53.0 && 81.3 & 76.4 \\
			\rowcolor{maroon}L2-loss & \underline{70.8} & \underline{53.7} && \underline{82.0} & \underline{76.6} \\
			\toprule[0.75pt]
		\end{tabular}
	}
\end{table}

% ------------ Regularization parameter $\lambda$ of regression loss ------
\begin{table}[!t]
	\centering
	\caption{Different $\lambda$ of regression loss (Smooth L1).}
	\label{table:abl_lambda}
	\setlength{\tabcolsep}{3.5mm}{
		\begin{tabular}{c c c c c c}
			\toprule[0.75pt]
			\multirow{2}{*}{$\lambda$} & \multicolumn{2}{c}{MFNet Dataset} & & \multicolumn{2}{c}{PST900 Dataset} \\
			\cmidrule[0.5pt]{2-3} \cmidrule[0.5pt]{5-6}
			& mAcc$\uparrow$ & mIoU$\uparrow$ & & mAcc$\uparrow$ & mIoU$\uparrow$ \\ 
			\midrule[0.5pt]
			\rowcolor{maroon}0.1 & 70.4 & 53.1 && 81.9 & 77.2 \\
			0.2 & 70.5 & 53.4 && 81.3 & 76.6 \\
			\rowcolor{maroon} 0.3 & \textbf{71.4} & \textbf{54.0} && \textbf{83.2} & \textbf{77.5} \\
			0.4 & \underline{71.0} & \underline{53.5} && \underline{82.0} & \underline{77.3} \\
			\rowcolor{maroon}0.5 & 70.5 & \underline{53.5} && 81.6 & 77.1\\
			\toprule[0.75pt]
		\end{tabular}
	}
\end{table}

\textbf{Dual-Encoder}. We adopt nine different encoder pairs to explore the suitable encoders for learning RGB and thermal features, and the results are shown in Table~\ref{table:abl_encoder}. The examined backbones involve ResNet \cite{he-cvpr2016-resnet}, DenseNet \cite{huang-cvpr2017-densenet}, and VGG \cite{simonyan-arxiv2014-vgg}. From the table, we see that asymmetric encoders achieves comparable with symmetric ones, or even better performance, at lower computational costs.  

% -------------- Dual encoder  -----------
\begin{table}[!t]
	\centering
	\caption{Ablation of the asymmetric dual-encoders.}
	\label{table:abl_encoder}
	%\resizebox{0.5\textwidth}{!}{}
	\setlength{\tabcolsep}{0.6mm}{
		\begin{tabular}{l c c c c c c c c}
			\toprule[0.75pt]
			\multirow{2}{*}{Backbone} & \multicolumn{3}{c}{MFNet Dataset} & &
			\multicolumn{3}{c}{PST900 Dataset} & \multirow{2}{*}{$\#$Params} \\
			\cmidrule[0.5pt]{2-4} \cmidrule[0.5pt]{6-8}
			& mAcc$\uparrow$ & mIoU$\uparrow$ & FLOPs(G)$\downarrow$ & & mAcc$\uparrow$ & mIoU$\uparrow$ & FLOPs(G)$\downarrow$ & (M)$\downarrow$ \\
			\midrule[0.5pt]
			\rowcolor{maroon}ResNet50-18 & 71.4 & 54.0 & \textbf{64.69} && 83.2 & 77.5 & \textbf{194.56} & 44.53 \\
			ResNet50-34 & 71.5 & 54.1 & \underline{76.03} && 83.3 & 77.6 & \underline{228.67} & 54.64 \\
			\rowcolor{maroon}ResNet50-50 & 71.7 & 54.3 & 79.55 && 83.5 & 77.7 & 239.26 & 57.71 \\
			%ResNet18-50 & 66.8 & 51.3 & \textbf{64.69} && 78.6 & 73.8 & \textbf{194.56} & 44.53 \\
			{\tiny DenseNet161-121} & 72.8 & 54.5 & 95.02 && 84.8 & 78.7 & 284.75 & \underline{44.10} \\
			\rowcolor{maroon}{\tiny DenseNet161-161} & 73.0 & 54.7 & 125.88 && 85.0 & 78.8 & 377.21 & 64.39 \\
			VGG19-16 & 69.1 & 53.1 & 241.61 && 81.0 & 75.2 & 724.86 & \textbf{43.91} \\
			\rowcolor{maroon}VGG19-19 & 69.4 & 53.3 & 267.10 && 81.3 & 75.5 & 801.32 & 49.22 \\
			{\scriptsize ResNet101-34} & \underline{73.3} & \underline{54.9} & 98.82 && \underline{85.5} & \underline{79.0} & 297.04 & 73.63\\
			\rowcolor{maroon}{\tiny ResNet101-101} & \textbf{73.6} & \textbf{55.1} & 125.13 && \textbf{85.7} & \textbf{79.1} & 376.00 & 95.70 \\
			\toprule[0.75pt]
		\end{tabular}
	}
\end{table}

% -------- Single encoder ------
\begin{table*}[!t]
	\centering
	\caption{Ablation of single encoder. NTE: RSFNet with No Thermal Encoder; NRE: RSFNet with No RGB Encoder.}
	\label{table:abl_singleEncoder}
	\setlength{\tabcolsep}{0.8mm}{
		\begin{tabular}{l c c c c c c c c c c c c c c c c c}
			\toprule[0.75pt]
			\multirow{3}{*}{Variants} & \multicolumn{8}{c} {MFNet Dataset} & & \multicolumn{8}{c} {PST900 Dataset}\\
			\cmidrule[0.5pt]{2-9} \cmidrule[0.5pt]{11-18}
			& \multicolumn{2}{c}{Ours} & & \multicolumn{2}{c}{NTE} & & \multicolumn{2}{c}{NRE} & & \multicolumn{2}{c}{Ours} & & \multicolumn{2}{c}{NTE} & & \multicolumn{2}{c}{NRE} \\
			\cmidrule[0.5pt]{2-3} \cmidrule[0.5pt]{5-6} \cmidrule[0.5pt]{8-9} \cmidrule[0.5pt]{11-12} \cmidrule[0.5pt]{14-15} \cmidrule[0.5pt]{17-18}
			& mAcc$\uparrow$ & mIoU$\uparrow$ & & mAcc$\uparrow$ & mIoU$\uparrow$ & & mAcc$\uparrow$ & mIoU$\uparrow$ & & mAcc$\uparrow$ & mIoU$\uparrow$ & & mAcc$\uparrow$ & mIoU$\uparrow$ & & mAcc$\uparrow$ & mIoU$\uparrow$\\
			\midrule[0.5pt]
			\rowcolor{maroon}RSFNet50-18 & 71.4 & 54.0 & & 65.7 & 49.7 & & 64.5 & 44.8 && 83.2 & 77.5 && 80.1 & 75.0 && 39.1 & 36.7 \\
			RSFNet161-121 & \underline{72.8} & \underline{54.5} & & \underline{66.9} & \underline{50.7} & & \underline{65.9} & \underline{45.7} & & \underline{84.8} & \underline{78.7} && \underline{83.0} & \underline{77.4} && \underline{45.2} & \underline{42.4} \\
			\rowcolor{maroon}RSFNet101-34 & \textbf{73.3} & \textbf{54.9} & & \textbf{68.9} & \textbf{51.3} & & \textbf{67.4} & \textbf{46.4} && \textbf{85.5} & \textbf{79.0} && \textbf{83.9} & \textbf{77.9} && \textbf{47.2} & \textbf{44.2} \\
			\toprule[0.75pt]
		\end{tabular}
	}
\end{table*}

\textbf{Single-Encoder}. We examine the performance of using single encoder for the RGB branch and the thermal branch independently, \ie, keep the one encoder while removing the other, and the results are shown in Table~\ref{table:abl_singleEncoder}. As shown in the table, abandoning the arbitrary branch, the performance drops greatly by about 7-8\%, which demonstrates the importance of the two-stream encoder that employs the two compensating modality features. In addition, the RGB encoder works better than the thermal encoder, which might be the reason that RGB images are taken under good lighting conditions and the objects are more easily to be identified.

\subsection{Qualitative Results}
\begin{figure}[!t]
	\centering
	\includegraphics[width=0.5\textwidth]{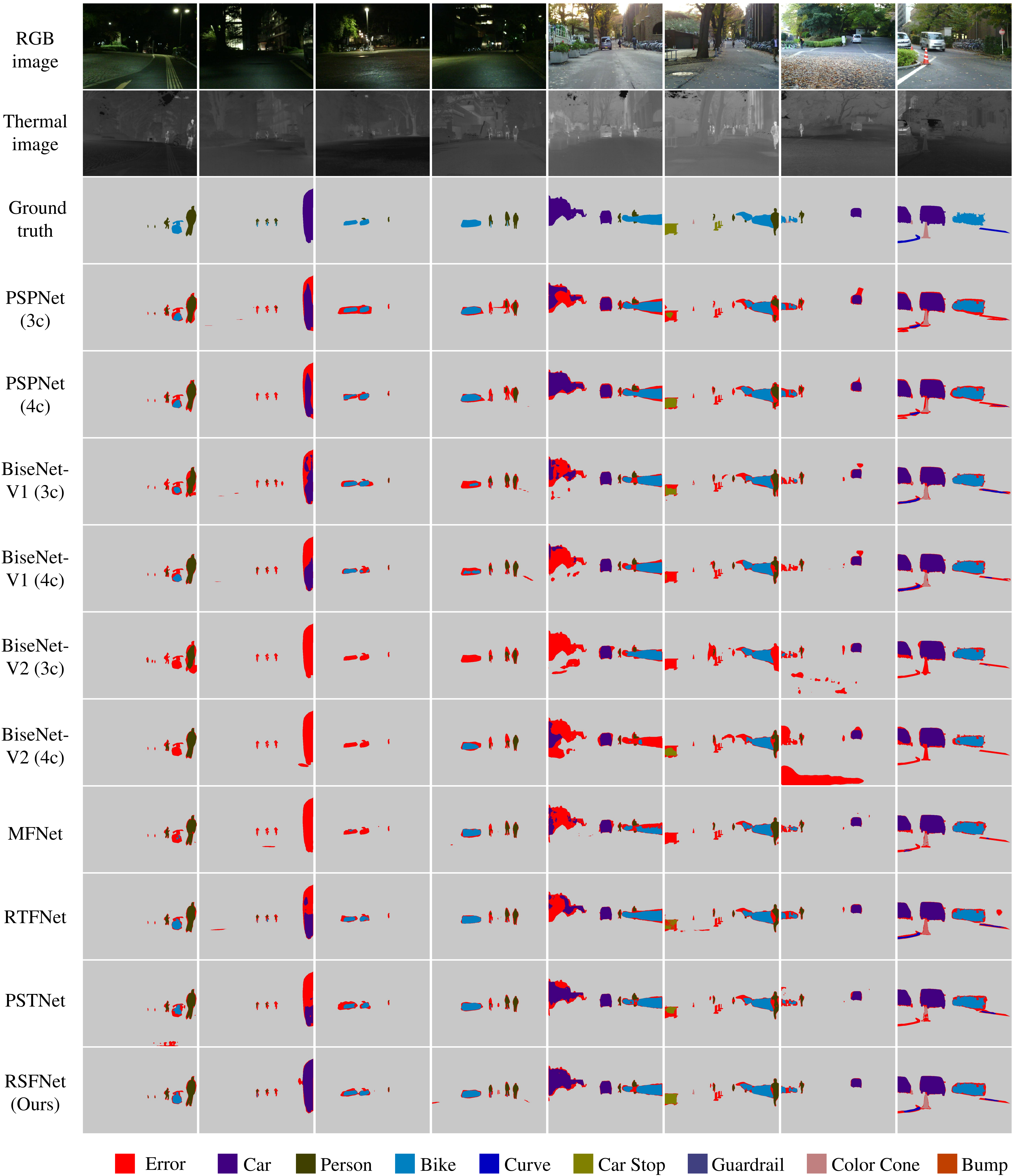}
	\caption{Segmentation results on MFNet test set. The left four columns use the nighttime images and the right four columns use the daytime images.}
	\label{fig:mfnet_visual}
\end{figure}

\begin{figure}[!t]
	\centering
	\includegraphics[width=0.5\textwidth]{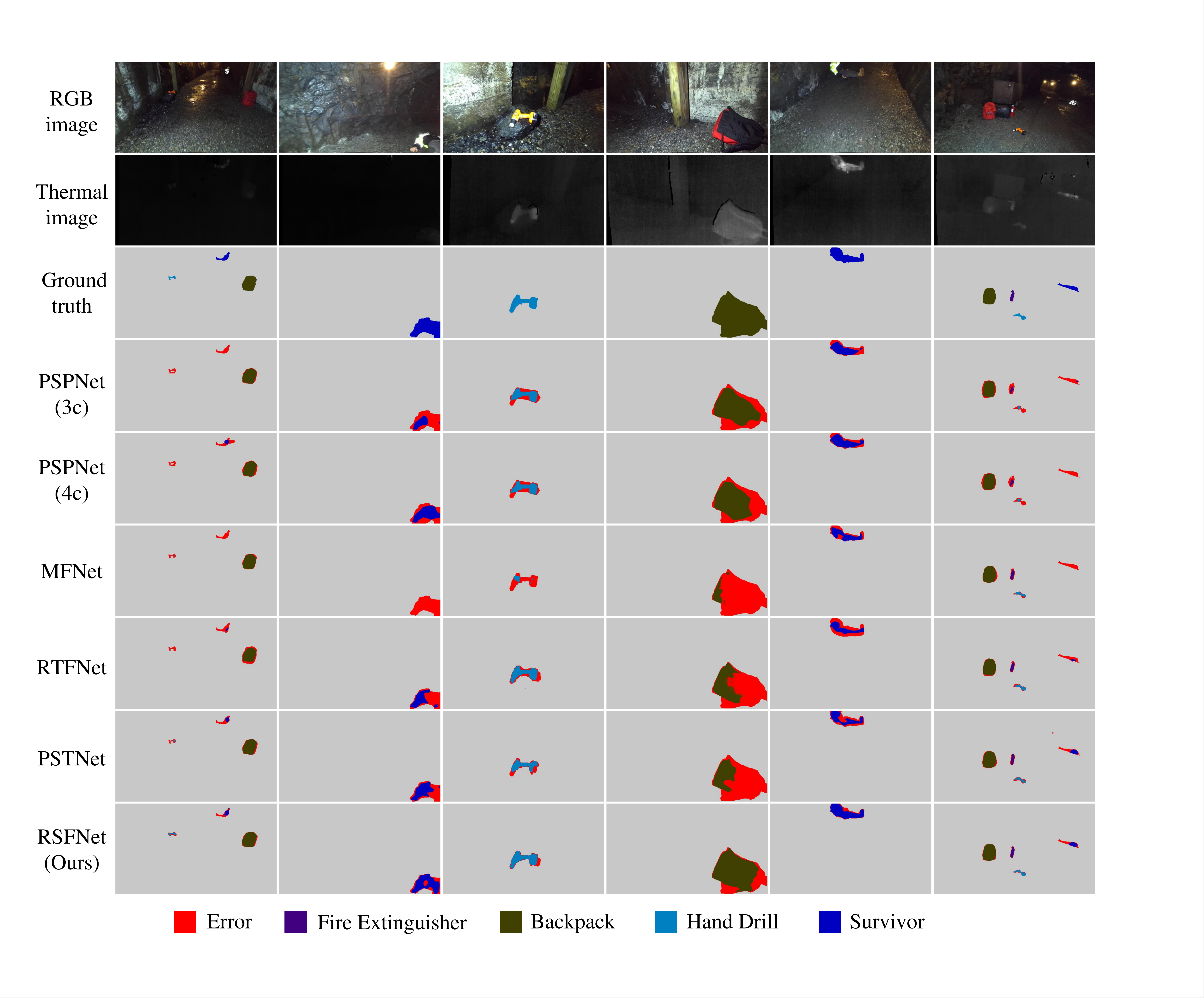}
	\caption{Segmentation results on PST900 test set.}
	\label{fig:pst900_visual}
\end{figure}

We randomly chose some images from the MFNet and the PST900 test set, and visualize their qualitative results in Fig.~\ref{fig:mfnet_visual} and Fig.~\ref{fig:pst900_visual}, respectively. From the figures, we observe that our RSFNet approach achieves the most satisfying segmentation results among the compared methods with the least error areas. For example, there are three persons in the 4-th nighttime image, which are difficult to identify from the RGB image ($row1$) in the dark environment, but they can be easily found in the thermal image ($row2$). This desires adopting the cross-modality fusion strategy to improve the segmentation performance. Compared to PSPNet \cite{zhao-cvpr2017-pspnet}, BiseNetV1 \cite{yu-eccv2018-bisenet}, BiseNetV2 \cite{yu-ijcv2021-bisenetv2}, MFNet \cite{ha-iros2017-mfnet}, RTFNet \cite{sun-ral2019-rtfnet}, and PSTNet \cite{shivakumar-icra2020-pst900}, our RSFNet method is able to segment the semantic areas more accurately in both daytime and nighttime scenes, which once again validates the effectiveness of our approach.

% --- Visualization on MFNet test set ------

% --- Visualization on PST900 test set ------

\section{Conclusion}
\label{conclusion}

This work has explored the cross-modality feature fusion strategy for RGB-Thermal semantic segmentation, and proposed the Residual Spatial Fusion Network (RSFNet). It adopts the asymmetric dual-encoder, and uses the saliency detection skill to generate the pseudo-label for supervising the feature learning, such that the saliency-aware RGB and thermal features can be learned. More importantly, we develop the residual spatial fusion module to integrate the dual-modality features using cross fusion along the spatial dimension. Both quantitative and qualitative experiments have been conducted on two benchmarks, and the results demonstrate the superiority of our method over the SOTA alternatives in the daytime, nighttime, and underground scenes.

%\bibliographystyle{IEEEtran}
%\bibliography{rgbt_seg_short}

\ifCLASSOPTIONcaptionsoff
  \newpage
\fi

\end{document}